\documentclass{article}

\usepackage[preprint]{neurips_2026}

\usepackage[utf8]{inputenc} 
\usepackage[T1]{fontenc}    
\usepackage{hyperref}       
\usepackage{url}            
\usepackage{booktabs}       
\usepackage{amsfonts}       
\usepackage{nicefrac}       
\usepackage{microtype}      
\usepackage{xcolor}         

\title{Intrinsically Interpretable Attention via Sparse Post-Training}

\usepackage{amsmath}
\usepackage{amssymb}
\usepackage{mathtools}
\usepackage{amsthm}
\usepackage{soul}
\usepackage{xcolor}
\usepackage[most]{tcolorbox}
\usepackage{subcaption}
\usepackage{wrapfig}

\definecolor{maintext}{RGB}{96,115,140}
\definecolor{highlight}{RGB}{255,243,191}
\definecolor{featureborder}{RGB}{180,190,205}
\definecolor{featurebg}{RGB}{245,247,250}
\definecolor{featuretitle}{RGB}{70,90,120}
\definecolor{bauhausblue}{HTML}{0061AC}
\definecolor{bauhausred}{HTML}{DF232C}
\definecolor{bauhausyellow}{HTML}{F1B50E}

\usepackage{listings}
\newtcolorbox{featurebox}[3]{%
  enhanced,
  colback=featurebg,
  colframe=featureborder,
  boxrule=0.8pt,
  arc=4pt,
  outer arc=4pt,
  left=10pt,
  right=10pt,
  top=8pt,
  bottom=8pt,
  title=\textbf{Feature #1} \hfill {\small #2},
  fonttitle=\color{featuretitle},
  coltitle=featuretitle,
  attach title to upper,
  before skip=14pt,
  after skip=14pt,
}
\usepackage[capitalize,noabbrev]{cleveref}

\author{%
Florent Draye\textsuperscript{1,*},
Anson Lei\textsuperscript{1,2,*},
Hsiao-Ru Pan\textsuperscript{1},
Ingmar Posner\textsuperscript{2},
Bernhard Schölkopf\textsuperscript{1,3}
\\
\textsuperscript{1}MPI-IS \quad
\textsuperscript{2}University of Oxford \quad
\textsuperscript{3}ETH Zürich \quad
\textsuperscript{*}Equal contribution
\\
\texttt{fdraye@tuebingen.mpg.de}
}

\begin{document}

\maketitle

\begin{abstract}
We introduce a simple post-training method that makes transformer attention sparse without sacrificing performance. Applying a flexible sparsity regularisation under a constrained-loss objective, we show on models up to 7B parameters that it is possible to retain the original pretraining loss while reducing attention connectivity to $\approx 0.4 \%$ of its edges. Unlike sparse-attention methods designed for computational efficiency, our approach leverages sparsity as a structural prior: it preserves capability while exposing a more organized and interpretable connectivity pattern. We find that this local sparsity cascades into global circuit simplification: task-specific circuits involve far fewer components (attention heads and MLPs) with up to 100× fewer edges connecting them. Additionally, using cross-layer transcoders, we show that sparse attention substantially simplifies attention attribution, enabling a unified view of feature-based and circuit-based perspectives. These results demonstrate that transformer attention can be made orders of magnitude sparser, suggesting that much of its computation is redundant and that sparsity may serve as a guiding principle for more structured and interpretable models.
\end{abstract}

\section{Introduction}
Scaling has driven major advances in artificial intelligence, with ever-larger models trained on internet-scale datasets achieving remarkable capabilities across domains. Large language models (LLMs) now underpin applications from text generation to question answering, yet their increasing complexity renders their internal mechanisms largely opaque~\citep{Bommasani2021FoundationModels}. Methods of mechanistic interpretability have been developed to address this gap by reverse-engineering neural networks to uncover how internal components implement specific computations and behaviors. Recent advances in this area have successfully identified interpretable circuits, features, and algorithms within LLMs~\citep{nanda2023progress, olsson2022context}, showing that large complex models can, in part, be understood mechanistically, opening avenues for improving transparency, reliability, and alignment~\citep{bereska2024mechanistic}.

\begin{figure}
    \centering
    \vspace{-10pt}

    \includegraphics[width=\linewidth]{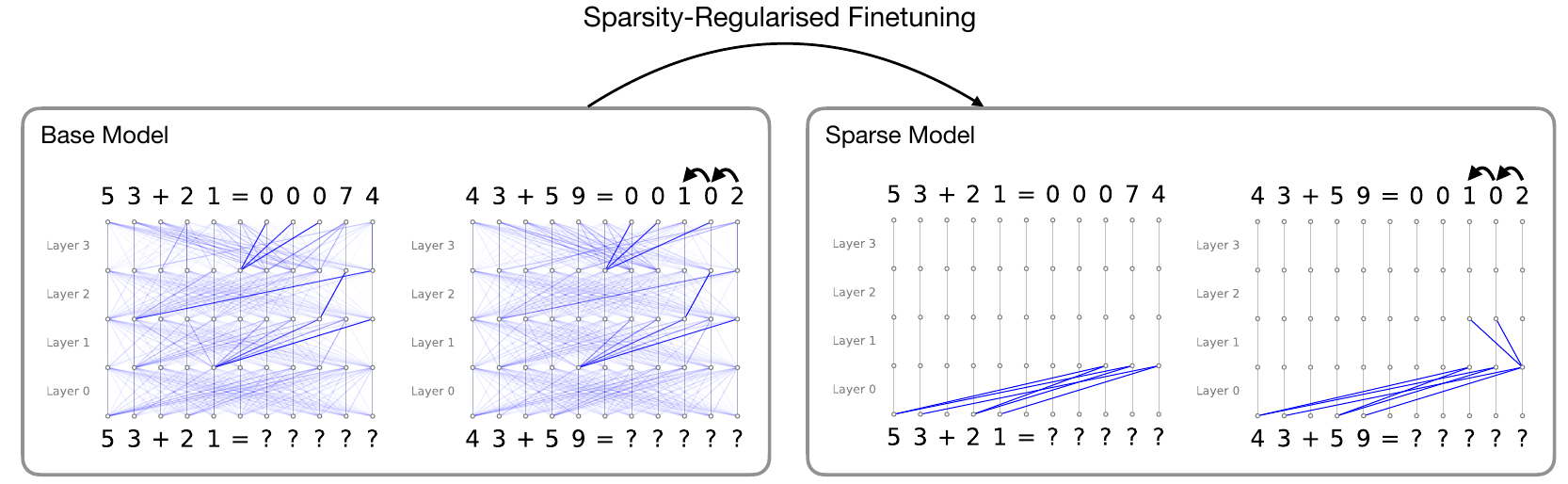}

    \caption{Visualised attention patterns for a 4-layer toy model trained on a simple 2-digit addition task. The main idea of this work is to induce \textit{sparse} attention between tokens via a post-training procedure that optimizes for attention sparsity while maintaining model performance. In this example, while both models are able to correctly predict the sum, the sparse model solves the problem with a naturally \textit{interpretable} circuit. Details of this toy setup and more examples are provided in Appendix~\ref{app:toy_model}.}
    \label{fig:open_illustration}

    \vspace{-10pt}
\end{figure}

However, interpretability is \textit{bottlenecked} by the model itself: even with sophisticated reverse-engineering techniques that can faithfully reveal internal algorithms, the underlying computations implemented by large models can still remain highly complex and uninterpretable. Circuits for seemingly simple tasks may span hundreds of interacting attention heads and MLPs with densely intertwined contributions across layers~\citep{conmy2023towards}, and features can influence each other along combinatorially many attention-mediated paths, complicating attention attribution~\citep{kamath2025tracing}. To exemplify this, Figure~\ref{fig:open_illustration} (top) illustrates the attention patterns of a small, single-head transformer trained on a simple two-digit addition task. Here, the model has learned to solve the task in a highly diffused manner, where information about each token is dispersed across all token locations, rendering the interpretation of the underlying algorithm extremely difficult even in this simple case.

The crux of the problem is that models are not incentivised to employ simple algorithms during training. In this work, we advocate for directly embedding interpretability constraints into model design in a way that induces simple circuits while preserving performance. 
We focus our analysis on attention mechanisms and investigate sparsity regularisation on attention patterns as an inductive bias.
To demonstrate how sparse attention patterns can give rise to interpretable circuits, we return to the two-digit addition example: Figure~\ref{fig:open_illustration} (bottom) shows the attention patterns induced by penalising attention edges during training. Here, the sparsity inductive bias forces the model to solve the problem with much smaller, intrinsically interpretable computation circuits.

In this work, we investigate using this sparsity regularisation scheme as a \textit{post-training} strategy for pre-trained LLMs. We propose a practical method for fine-tuning existing models without re-running pretraining, offering a flexible way to induce sparse attention patterns and enhance interpretability. We show, on models of up to 7B parameters, that our proposed procedure preserves the performance of the base models on pretraining data while reducing the effective attention map to less than $0.4\%$ of its edges.
To evaluate our central hypothesis that sparse attention facilitates interpretability, we consider two complementary settings.
First, we study circuit discovery, where the objective is to identify the minimal set of components responsible for task performance~\citep{conmy2023towards}. We find that sparsified models yield substantially simpler computational graphs: the resulting circuits explain model behaviour using up to four times fewer attention heads and up to two orders of magnitude fewer edges.
Second, using cross-layer transcoders~\citep{ameisen2025circuit}, we analyse attribution graphs, which capture feature-level interactions across layers. In this setting, sparse attention mitigates the attention attribution problem by making it possible to identify which attention heads give rise to a given edge, owing to the reduced number of components mediating each connection. We argue that this clarity enables a tighter integration of feature-based and circuit-based perspectives, allowing feature interactions to be understood through explicit, tractable circuits.
Taken together, these results position attention sparsity as an effective and practical inductive tool for surfacing the minimal functional backbone underlying model behaviour.

\section{Related Work}

\subsection{Sparse Attention}
As self-attention is a key component of the ubiquitous Transformer architecture, a large number of variants of attention mechanisms have been explored in the literature. 
Related to our approach are sparse attention methods, which are primarily designed to alleviate the quadratic scaling of vanilla self-attention.
These methods typically rely on masks based on fixed local and strided patterns~\citep{child2019generating} or sliding-window and global attention patterns~\citep{beltagy2020longformer, zaheer2020big} to constrain the receptive field of each token. 
While these approaches are successful in reducing the computational complexity of self-attention, they require hand-defined heuristics that do not reflect the internal computations learned by the model.

Beyond these fixed-pattern sparse attention methods, Top-$k$ attention, which enforces sparsity by dynamically selecting the $k$ most relevant keys per query based on their attention scores, has also been explored~\citep{gupta2021memory, deepseekai2025deepseekv32pushingfrontieropen}. 
While Top-$k$ attention enables \textit{learnable} sparse attention, it is poorly suited as a post-training interpretability intervention. 
After training, imposing a hard Top-$k$ constraint directly breaks model performance. 
The choice of $k$ is also brittle: smaller values remove necessary edges, whereas larger values leave attention patterns too dense to interpret. 
In contrast, our L1 regularisation induces sparsity gradually during training, preserving performance while avoiding the need to manually choose a discrete sparsity level $k$. 
Most importantly, Top-$k$ attention imposes the same number of active edges for every query, rather than allowing sparsity to vary across heads and contexts. This flexibility is crucial in the sparse regimes we study: for instance, in our GPT-2 setting, only $0.22\%$ of edges remain active, yielding far fewer than one active edge per query on average, a regime that cannot be captured by Top-$k$ attention even with $k=1$. Recently, \citet{lei2025spartan} introduce a sparsity regularisation scheme for world modelling that reveals sparse token dependencies. We build on this method and examine its role as an inductive bias for interpretability.

\subsection{Circuit Discovery and Attribution Graphs}


Mechanistic interpretability seeks to uncover how internal components of LLMs implement specific computations. Ablation studies assess performance drops from removing components~\citep{nanda2023progress}, activation patching measures the effect of substituting activations~\citep{zhang2023towards}, and attribution patching scales this approach via local linearisation~\citep{syed2024attribution}. Together, these approaches allow researchers to isolate sub-circuits, minimal sets of attention heads and MLPs that are causally responsible for a given behavior or task~\citep{conmy2023towards}. Attention itself plays a dual role: it both routes information and exposes interpretable relational structure, making it a key substrate for mechanistic study. Our work builds on this foundation by leveraging sparsity to simplify these circuits, amplifying the interpretability of attention-mediated computation while preserving model performance.


Mechanistic interpretability has gradually shifted from an emphasis on explicit circuit discovery towards the analysis of internal representations and features. Recent work on attribution graphs and circuit tracing seeks to reunify these perspectives by approximating MLP outputs as sparse linear combinations of features and computing causal effects along linear paths between them~\citep{dunefsky2024transcoders, ameisen2025circuit, lindsey2025biology}. This framework enables the construction of feature-level circuits spanning the computation from input embeddings to final token predictions. Within attribution graphs, edges correspond to direct linear causal relationships between features. However, these relationships are mediated by attention heads that transmit information across token positions. Identifying which attention heads give rise to a particular edge, and understanding why they do so, is essential, as this mechanism forms a fundamental component of the computational graph~\citep{kamath2025tracing}.
A key limitation of current attribution-based approaches is that individual causal edges are modulated by dozens of attention components. We show that this leads to feature-to-feature influences that are overly complex, rendering explanations in terms of other features in the graph both computationally expensive and conceptually challenging. 


\section{Method}
Our main hypothesis is that post-training existing LLMs to encourage sparse attention patterns leads to the emergence of more interpretable circuits.
In order to instantiate this idea, we require a post-training pipeline that satisfies three main desiderata: 
\begin{enumerate}
    \item To induce sparse message passing between tokens, we need an attention mechanism that can `zero-out' attention edges, which in turn enables effective $L_0$-regularisation on the attention weights. This is in contrast to the standard softmax attention mechanism, where naive regularisation would result in small but non-zero attention weights that still allow information flow between tokens. 
    \item The model architecture needs to be compatible with the original LLM such that the pre-trained LLM weights can be directly loaded at initialisation.
    \item The post-training procedure needs to ensure that the post-trained models do not lose prediction performance compared to their fully-connected counterparts.
\end{enumerate}
To this end, we leverage the Sparse Transformer architecture in the SPARTAN framework proposed in \cite{lei2025spartan}, which uses sparsity-regularised \textit{hard attention} instead of the standard softmax attention.
In the following subsections, we describe the Sparse Transformer architecture and the optimisation setup, highlighting how this approach satisfies the above desiderata.

\subsection{Sparse Attention Layer}
Given a set of token embeddings, the Sparse Transformer layer computes the key, query, and value embeddings, $\{k_i, q_i, v_i\}$, via linear projections, analogous to the standard Transformer. Based on the embeddings, we sample a binary gating matrix from a learnable distribution parameterised by the keys and queries,
\begin{equation}
    A_{ij} \sim  \mathrm{Bern}(\sigma(q_i^T k_j)),
\end{equation}
where $\mathrm{Bern}(\cdot)$ is the Bernoulli distribution and $\sigma(\cdot)$ is the logistic sigmoid function.
This sampling step can be made differentiable via the Gumbel Softmax trick~\citep{jang2017categorical}.
This binary matrix acts as a mask that controls the information flow across tokens. 
Next, the message passing step is carried out in the same way as standard softmax attention, with the exception that we mask out the value embeddings using the sampled binary mask,
\begin{equation}
    \label{eqn:sparse_attn}
    \mathrm{SparseAttn}(Q, K, V) = \bigg [A \odot \mathrm{softmax}(\frac{QK^T}{\sqrt{d_k}})\bigg]V,
\end{equation}
where $d_k$ is the dimension of the key embeddings and $\odot$ denotes element-wise multiplication.
During training, we regularise the expected number of edges between tokens based on the distribution over the gating matrix. Concretely, the expected number of edges for each layer can be calculated as
\begin{equation}
    \mathbb{E}\big[|A|\big] = \sum_{i,j} \sigma(q^T_ik_j).
\end{equation}
Note that during the forward pass, each entry of $A$ is a \textit{hard} binary sample that zeros out attention edges, which serves as an effective $L_0$ regularisation.
Moreover, since the functional form of the sparse attention layer after the hard sampling step is the same as standard softmax attention, pre-trained model weights can be directly used without alterations.\footnote{Technically, the sampled $A$ affects the computation. This can be mitigated by adding a positive bias term inside the sigmoid function to ensure all gates are open at initialisation. Experimentally, we found this to be unnecessary as the models quickly recover their original performance within a small number of gradient steps.}

\subsection{Constrained Optimisation}
In order to ensure that the models do not lose prediction performance during the post-training procedure, as per desideratum 3, we follow the approach proposed by \citet{lei2025spartan}, which employs the GECO algorithm~\citep{rezende2018tamingvaes}.
Originally developed in the context of regularising VAEs, the GECO algorithm places a constraint on the performance of the model and uses a Lagrangian multiplier to automatically find the right strength of regularisation during training. 
Concretely, we formulate the learning process as the following optimisation problem,
\begin{equation}
    \min_\theta \sum_l \mathbb{E}\big[|A_l|\big] \qquad  s.t. \quad CE \leq \tau,
\end{equation}
where $A_l$ denotes the gating matrix at layer $l$, $CE$ is the standard next token prediction cross-entropy loss, and $\tau$ is the required target loss, and $\theta$ is the model parameters. In practice, we set this target as the loss of the pre-trained baseline models.
We solve this optimisation problem via Lagrangian relaxation, yielding the following max-min objective,
\begin{equation}
    \max_{\lambda>0}\min_\theta \bigg [ \sum_l \mathbb{E}\big[|A_l|\big] + \lambda(CE - \tau) \bigg].
\end{equation}
This can be solved by taking gradient steps on $\theta$ and $\lambda$ alternately.
During training, updating $\lambda$ automatically balances the strength of the sparsity regularisation: when $CE$ is lower than the threshold, $\lambda$ decreases, and hence more weight is given to the sparsity regularisation term. This effectively acts as an adaptive schedule which continues to increase the strength of the regularisation until the model performance degrades. Here, the value of $\tau$ is selected as a hyperparameter to ensure that the sparse model's performance remains within a certain tolerance of the original base model. In practice, the choice of $\tau$ controls a trade off between sparsity and performance: picking a tight $\tau$ can lead to a slower training process, whereas a higher tolerance can substantially speed up training at the cost of potentially harming model performance. In Appendix~\ref{app:training}, we provide further discussion on this optimisation process and its training dynamics.

\subsection{Practical Considerations}
One of the main strengths of our proposed method is that, architecturally, the only difference between a sparse Transformer and a normal one lies in how the dot-product attention is computed. As such, most practical training techniques for optimising Transformers can be readily adapted to our setting. In our experiments, we find the following techniques helpful for improving computational efficiency and training stability.

\begin{itemize}
    \item \textbf{FlashAttention}~\citep{dao2023flashattention}. FlashAttention has become a standard method for reducing the GPU memory footprint of dot-product attention mechanisms. In Appendix~\ref{app:splash_attention}, we discuss how the sampled sparse attention can be implemented in an analogous manner.

    \item \textbf{LoRA finetuning}~\citep{hu2022lora}. Low rank finetuning techniques can significantly reduce the computational requirements for training large models. In our experiments, we verify on a 7B parameter model that LoRA finetuning is sufficiently expressive for inducing sparse attention patterns.

    \item \textbf{Distillation}~\citep{gu2024minillm}. Empirically, we find that adding an auxiliary distillation loss based on the KL divergence between the base model and the sparse model improves training stability and ensures that the behaviour of the model remains unchanged during post-training.
\end{itemize}
\section{Experiments}




To evaluate the effectiveness of our post-training pipeline, we finetune pre-trained LLMs and compare their prediction performance and interpretability before and after applying sparsity regularisation.
We perform full finetuning on a GPT-2 base model~\citep{radford2019language}(124M parameters) on the OpenWebText dataset~\citep{Gokaslan2019OpenWeb}. To investigate the generality and scalability of our method, we perform LoRA finetuning on the larger OLMo-7B model~\citep{Groeneveld2023OLMo} on the Dolma dataset~\citep{dolma}, which is the dataset on which the base model was trained. The GPT-2 model and the OLMo model are trained on sequences of length 64 and 512, respectively. 
In the following subsections, we first present a quantitative evaluation of model performance and sparsity after sparse post-training. We then conduct two interpretability studies, using activation patching and attribution graphs, to demonstrate that our method enables the discovery of substantially smaller circuits.

\subsection{Model Performance and Sparsity}


We begin by evaluating both performance retention and the degree of sparsity achieved by post-training. We set cross-entropy targets of 3.50 for GPT-2 (base model: 3.48) and 2.29 for OLMo (base model: 2.25). After training, the mean cross-entropy loss for both models remains within $\pm 0.01$ of the target, indicating that the dual optimisation scheme effectively enforces a tight performance constraint.
To quantify the sparsity achieved by the models, we evaluate them on the validation split of their respective datasets and compute the mean number of non-zero attention edges per attention head. We find that the sparsified GPT-2 model activates, on average, only \textbf{0.22\%} of its attention edges, while the sparsified OLMo model activates \textbf{0.44\%}, indicating substantial sparsification in both cases. 
To further verify that this drastic reduction in message passing between tokens does not substantially alter model behaviour, we evaluate the sparsified OLMo model on a subset of the benchmarks used to assess the original model. As shown in Figure~\ref{fig:olmo_benchmark}, the sparse model largely retains the performance of the base model across a diverse set of tasks. 
In sum, our results demonstrate that sparse post-training is effective in consolidating information flow into a small number of edges while maintaining a commensurate level of performance.

\subsection{Circuit Discovery with Activation Patching}
\label{sec:circuit_discovery}
\begin{figure}
    \centering
    \vspace{-10pt} 

    \includegraphics[width=\linewidth]{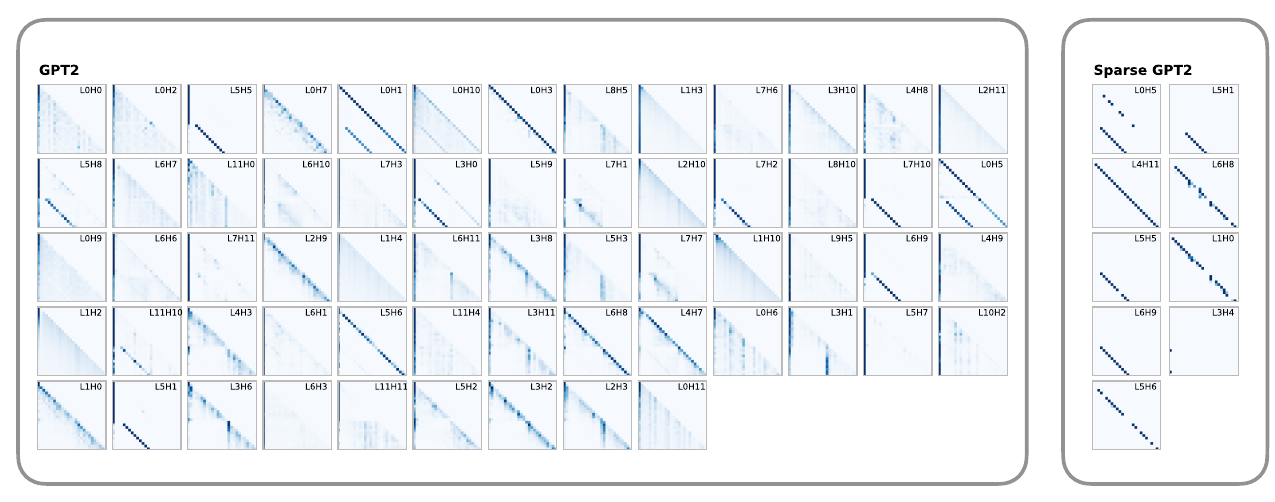}

    \caption{Attention patterns of the heads required to explain 90\% of model behaviour on a \textit{copy} task. The sparse model requires substantially fewer attention heads. Moreover, the selected heads exhibit the characteristic `induction head' pattern: each token attends to a previous token at a fixed relative offset, effectively copying information forward through the sequence, a pattern well known to implement the copy mechanism in transformer models. Equivalent plots for OLMo can be found in Appendix~\ref{app:extra_experiments}.}
    \label{fig:induction_head}

    \vspace{-10pt} 
\end{figure}



We begin by outlining the experimental procedure used for circuit discovery. \textit{Activation patching}~\citep{nanda2023progress} is a widely used technique for identifying task-specific circuits in transformer models. In a typical setup, the model is evaluated on pairs of prompts: a \textit{clean} prompt, for which the model predicts a correct target token, and a \textit{corrupted} prompt that shares the overall structure of the clean prompt but is modified to induce an incorrect prediction. Here, the goal is to find the set of model components that is responsible for the model's preference for the correct answer over the wrong one, as measured by the logit difference between the corresponding tokens. In activation patching, individual model components, such as attention heads and individual edges, can be 'switched-off' by patching activation at the specific positions. Circuit discovery amounts to finding a set of components whose replacement causes the model’s prediction to shift from the correct to the corrupted answer.

Since searching over every possible subset of model components is infeasible due to the exponential number of potential subsets, we adopt a common heuristic to rank each model component. Specifically, for each individual component, we compute an importance score by replacing the activations of the component with the corrupted activations and measuring its effect on the logit difference. In our experiments, we use this ranking to select the top-$k$ components and intervene on the model by freezing all remaining components, with the goal of identifying the minimal set that accounts for at least 90\% of the model’s preference for the correct prediction.
Note that these importance scores can be computed at two levels: (i) a \emph{single-sentence} level, using a single pair of correct and corrupted inputs, and (ii) a \emph{global} level, obtained by averaging scores across many task variants. In our experiments, we report the results using \textit{single-sentence} scores. In Appendix~\ref{app:extra_experiments}, we also provide results using the \textit{global} scores, which are largely consistent with our main results. There are also two standard approaches for freezing component activations: setting the activation to zero or replacing it with a mean activation value~\citep{conmy2023towards}. We evaluate both variants for each model and report results for the patching strategy that yields the smallest circuits.

We first focus on the \textit{copy} task with the following prompt: \texttt{"AJEFCKLMOPQRSTVWZS, AJEFCKLMOPQRSTVWZ"}, where the model has to copy the letter \texttt{S} to the next token position. This task is well studied and is widely believed to be implemented by emergent \textit{induction heads}~\citep{elhage2021mathematical}, which propagate token information forward in the sequence. Figure~\ref{fig:induction_head} illustrates the attention patterns of the set of attention heads that explains this prompt for the sparse and base GPT-2 models. See Appendix~\ref{app:extra_experiments} for analogous results for the OLMo models. The sparse model admits a substantially smaller set of attention heads (9 heads) than its fully connected counterpart (61 heads). Moreover, the identified heads in the sparse model exhibit cleaner induction head patterns, with each token attending to a single prior position at a fixed relative offset. These results illustrate how sparsification facilitates interpretability under simple ranking-based methods and support our hypothesis that sparse post-training yields models that are more amenable to mechanistic interpretability techniques.

To further verify our hypothesis, we repeat the experiment on classical circuit discovery tasks. For GPT-2, we evaluate variants of the \textit{Indirect Object Identification} (IOI) task, in which the model copies a person's name from the start of a sentence, and the \textit{Greater Than} task, in which the model predicts a number that is larger than a previously mentioned number. To further assess the scalability of our approach, we investigate more challenging and longer horizon tasks for OLMo, including a longer context IOI task and a \textit{Docstring} task where the model needs to predict an argument name in a Docstring based on an implemented function. Details of each task can be found in Appendix~\ref{app:circuit_discovery_tasks}.
Figure~\ref{fig:patching_heads_edges_single} shows the fraction of model behaviour explained as a function of the number of retained model components (attention heads and attention edges, respectively). Across all tasks and models, the sparse models consistently produce significantly smaller circuits, as measured by the number of model components needed to explain 90\% of model prediction. This further corroborates our claim that sparse models lead to simpler and more interpretable internal circuits. 

\begin{figure*}[t]
    \centering

    \includegraphics[width=\textwidth]{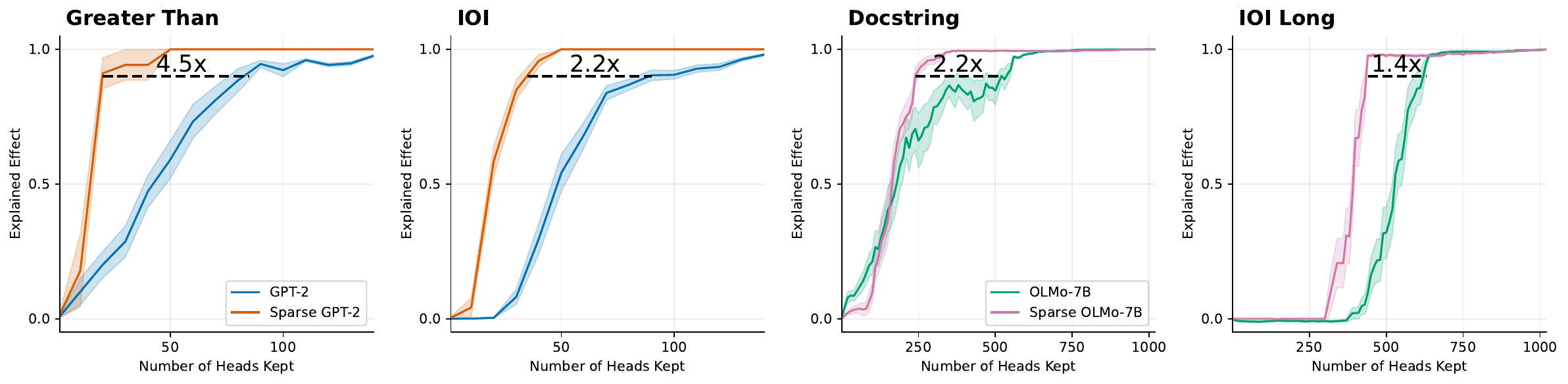}
    \vspace{0.3em}
    \includegraphics[width=\textwidth]{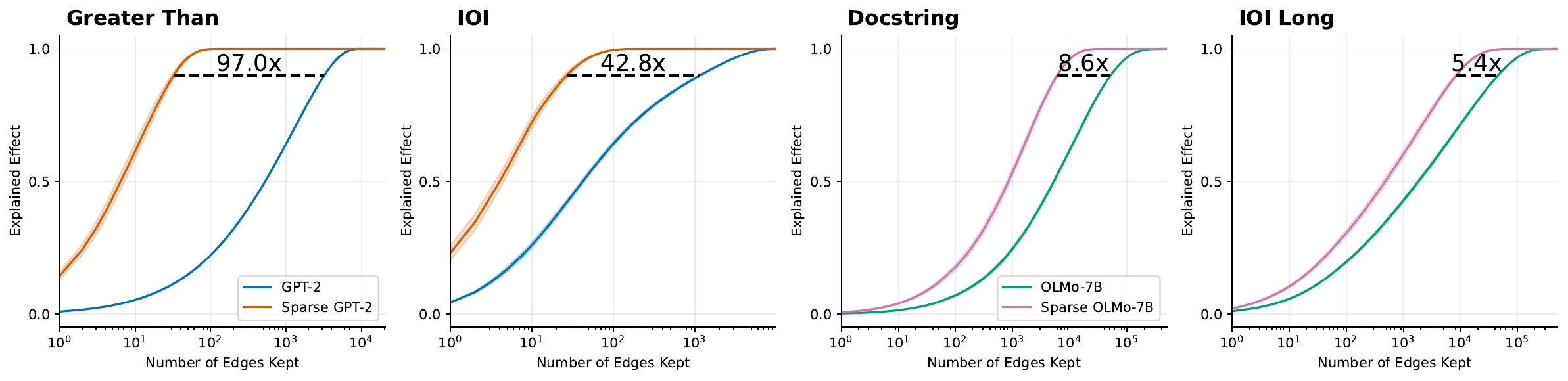}

    \caption{
    Logit attribution when keeping only the top-$k$ attention components.
    \textbf{Top:} attention \textbf{heads}. \textbf{Bottom:} attention \textbf{edges}.
    Dotted lines indicate the number of components required to explain 90\% of the
    logit difference. Sparse models yield 1.4$\times$--4.5$\times$ smaller circuits
    at the head level and 5.4$\times$--97$\times$ smaller circuits at the edge level.
    Shaded areas show standard error across 20 prompts.
    }
    \label{fig:patching_heads_edges_single}
\end{figure*}





\subsection{Attribution graphs}

\begin{wrapfigure}{r}{0.48\textwidth}
    \centering
    \vspace{-10pt} 

    \includegraphics[width=\linewidth]{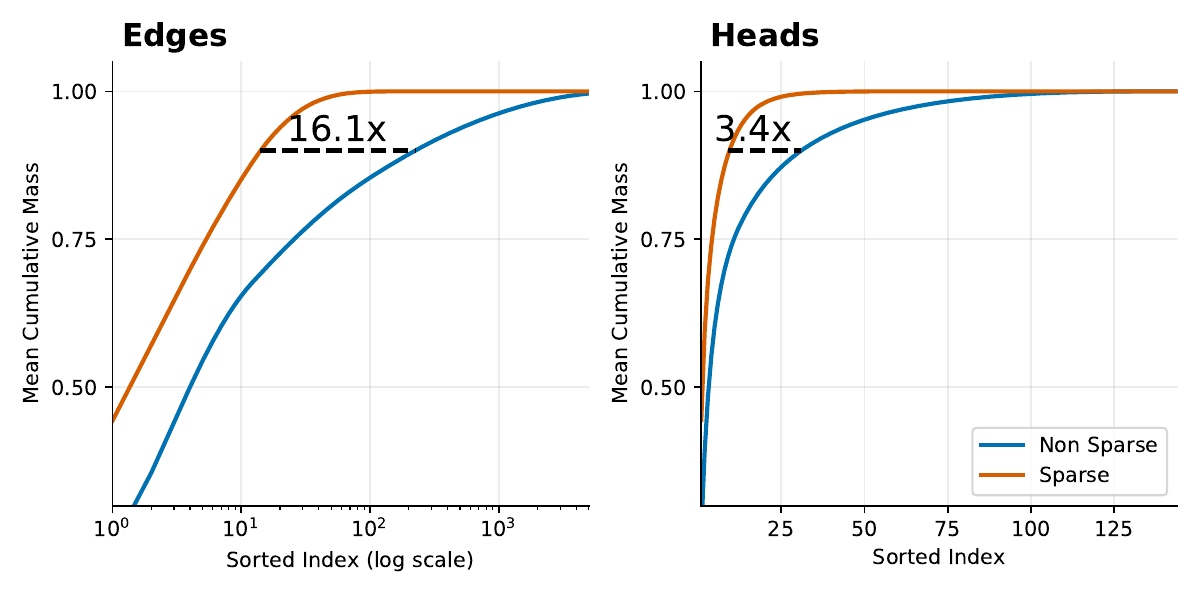}

    \caption{Mean cumulative distribution of the component scores that mediate an attribution graph edge. The components are on the left key-query pairs within a head, and on the right full attention heads.}
    \label{fig:mean_cum_sum}

    \vspace{-10pt} 
\end{wrapfigure}

Next, we present a more fine-grained, feature-level investigation of whether sparsity in attention leads to interpretable circuits in practice using cross-layer transcoders (CLTs). Since training CLTs on OLMo-7B is computationally prohibitive\footnote{The largest open-source CLT is on Gemma-2B at the time of writing.}, we focus our analysis on the GPT-2 models.
For the rest of the section, we perform analysis on CLTs trained on the sparse and base GPT-2 models, trained with an expansion factor of $32$ and achieve above $80\%$ replacement score measured with Circuit Tracer~\citep{circuit-tracer}. See Appendix~\ref{app:cross-layer-transcoder} and~\ref{app:attribution} for details on training and visualisation.

We study the problem of \textit{attention attribution}, which seeks to understand how edges between features are \textit{mediated}. The key challenge here is that any given edge can be affected by a large number of model components, making mediation circuits difficult to analyse both computationally and conceptually: computationally, exhaustive enumeration is costly; conceptually, the resulting circuits are often large and uninterpretable. In this experiment, we demonstrate that sparse attention patterns induced via post-training substantially alleviate these challenges, as the vast majority of attention components have zero effect on the computation.

As in \cite{ameisen2025circuit}, we define the total attribution score between feature $n$ at layer $\ell$ and position $k$, and feature $n'$ at layer $\ell'$ and position $k'$ as
\begin{equation}
a_{\ell, k, n}^{\ell', k', n'}
=
f_{k,n}^{\ell}
\;
J_{\ell, k}^{\ell', k'}
\;
g_{k',n'}^{\ell'} .
\end{equation}
Here, $f_{k,n}^{\ell}$ denotes the decoder vector corresponding to feature $n$ at layer $\ell$ and position $k$, and $g_{k',n'}^{\ell'}$ is the corresponding encoder vector for feature $n'$ at layer $\ell'$ and position $k'$. The term $J_{\ell, k}^{\ell', k'}$ is the Jacobian from the MLP output at $(\ell, k)$ to the MLP input at $(\ell', k')$. This Jacobian is computed during a forward pass in which all nonlinearities
are frozen using stop-gradient operations. Under this linearisation, the attribution score represents the sum over all linear paths from the source feature to the target feature.

To analyse how this total effect between two features is mediated by each model component, we define the component-specific attribution by subtracting the contribution of all paths that do not pass through the component:
\begin{equation*}
a_{\ell, k, n}^{\ell', k', n'}(h)
=
f_{k,n}^{\ell}
\;
J_{\ell, k}^{\ell', k'}
\;
g_{k',n'}^{\ell'}
-
f_{k,n}^{\ell}
\;
\bigl[J_{\ell, k}^{\ell', k'}\bigr]_h
\;
g_{k',n'}^{\ell'} .
\end{equation*}
Here, $\bigl[J_{\ell, k}^{\ell', k'}\bigr]_h$ denotes a modified Jacobian computed under the same linearization as above, but with the specific attention component $h$ additionally frozen via stop-gradient. As such, these component-specific scores quantifies how much each model component impacts a particular edge between features. 

\begin{figure*}[t]
    \includegraphics[width=1.\textwidth]{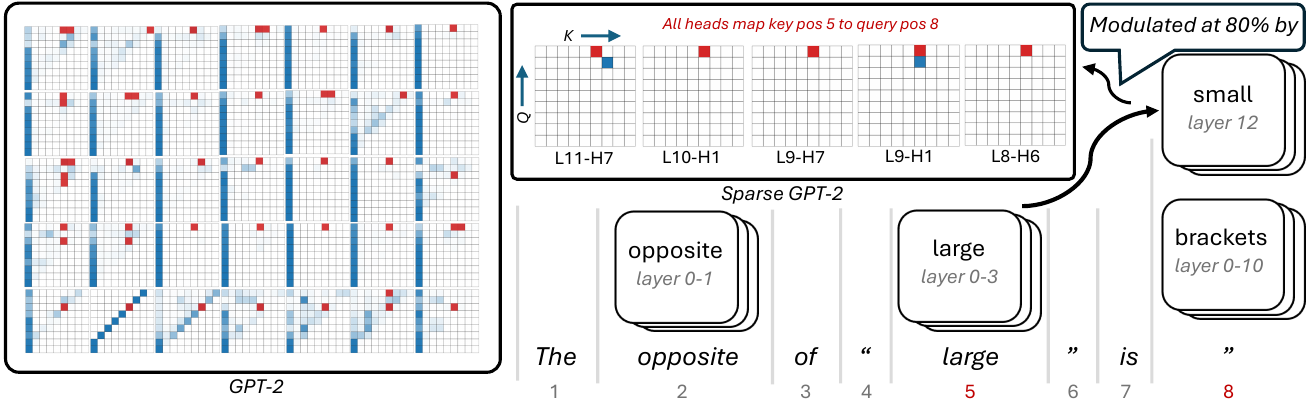}
    \caption{Sketch of the attribution graph for the sentence \emph{``The opposite of `large' is''}. The cluster of features associated with \emph{large} at token position 5 maps directly to the final next-token prediction logit \emph{small}. We show the attention patterns of all key--query pairs required to account for $80\%$ of the cumulative attribution score. In the sparse-attention setting, this corresponds to five attention heads, compared to more than forty heads in the dense-attention case. In the sparse model, these heads read from token position 5 and write directly to the last token residual stream at token position 8. These heads thus compute in parallel and provide a clear picture of the internal computation. }
    \label{fig:attribution_graph}
\end{figure*}

Empirically, we evaluate the method on ten pruned attribution graphs, computed on the IOI, greater-than, completion, and category tasks. Similar to our previous circuit discovery experiment, we compute attribution scores on the level of attention heads as well as individual key--query pairs. 
In practice, attention sparsity yields substantial computational savings: because inactive key--query pairs are known \emph{a priori} to have exactly zero attribution score, attribution need only be computed for a small subset of components. This reduces the computation time per attribution graph from several hours to several minutes.

In terms of circuit size, Figure~\ref{fig:mean_cum_sum} shows the mean cumulative distribution of component attribution scores for each edge in the attribution graph.
We find that, to reach a cumulative attribution threshold of $90\%$, the sparse model on average requires $16.1\times$ fewer key--query pairs and $3.4\times$ fewer attention heads when compared to the dense GPT-2 model, supporting our hypothesis that sparse attention patterns lead to simpler mediation circuits. 

Next, we present a qualitative case-study to showcase the benefits of sparse attention patterns.
For a given key--query pair, we compute the causal effect from all other features in the attribution graph to both the key and the query vectors. 
Figure~\ref{fig:attribution_graph} illustrates this analysis for the prompt \emph{``The opposite of `large' is''}. The resulting attribution graph decomposes into four coherent clusters of features: features related to \emph{opposite}, features related to \emph{large}, features activating on bracketed tokens, and the final next-token logit corresponding to \emph{small} (see Appendix~\ref{app:opposite_of} for examples of features and visualization).

Here, the features in the \emph{large} cluster are directly connected to the \emph{small} logit. The key question is then to understand how this connection from the \emph{large} to the \emph{small} logit comes about. To this end, we analyse their mediation structure. We find that $80\%$ of the cumulative absolute attribution score of the edges connecting the \emph{large} cluster to the \emph{small} logit is mediated by the same five late layer attention key--query pairs. These attention components map features from token position $5$ directly into the final-layer residual stream at position $8$, and thus operate in parallel. 

For these five key--query pairs, we then compute the causal influence of all other features in the graph on their key and query vectors. The query vectors are primarily modulated by features associated with bracketed tokens in the last token position, while the key vectors are driven by strongly active features in both the \emph{opposite} and \emph{large} clusters, details are provided in Appendix~\ref{app:opposite_of} and illustrated in Figure~\ref{fig:query_features}. These results are in agreement with the recent work on attention attribution and the "opposite of" attribution graph \citep{kamath2025tracing}. 

In stark contrast, Figure~\ref{fig:attribution_graph} (left) shows that a similar (and more computationally expensive) analysis on the dense model produces a much more complicated circuit. This case study illustrates the potential of sparse attention in the context of attribution graphs, as it enables a unified view of features and circuits. By jointly analyzing feature activations, attention components, and their mediating roles, we obtain a more faithful picture of the computational graph underlying the model’s input--output behavior.

\section{Conclusion}
Achieving interpretability requires innovations in both interpretation techniques \textit{and} model design. 
We investigate how large models can be trained to be intrinsically interpretable. 
We present a flexible post-training procedure that sparsifies transformer attention while preserving the original pretraining loss. By minimally adapting the architecture, we apply a sparsity penalty under a constrained-loss objective, allowing the pre-trained model to reorganise its connectivity into a much more selective and structured pattern. Mechanistically, this induced sparsity gives rise to substantially simpler circuits: task-relevant computation concentrates into a small number of attention heads and edges. Across a range of tasks and analyses, we show that sparsity improves interpretability at the \textit{circuit} level by reducing the number of components involved in specific behaviours. In circuit discovery experiments, most of the model’s behaviour can be explained by circuits that are orders of magnitude smaller than in dense models; in attribution graph analyses, the reduced number of mediating components renders attention attribution tractable. Together, these results position sparse post-training of attention as a practical and effective tool for enhancing the mechanistic interpretability of pre-trained models.

\paragraph{Limitations and Future Work.} 
One limitation of the present investigation is that, while we deliberately focus on sparsity as a post-training intervention, it remains an open question whether injecting a sparsity bias directly during training would yield qualitatively different or simpler circuit structures.  Moreover, our study is primarily restricted to sparsifying attention patterns, the underlying principle of leveraging sparsity to promote interpretability naturally extends to other components of the transformer architecture. As such, combining the proposed method with complementary approaches for training intrinsically interpretable models, such as Sparse Mixture-of-Experts~\citep{yang2025mixture}, sparsifying model weights~\citep{gao2024weightsparse}, or limiting superposition~\citep{miller2026identifyingintervenableinterpretablefeatures} offers a promising direction for future work.
Another exciting avenue for future work is to apply the sparsity regularisation framework developed here within alternative post-training paradigms, such as reinforcement learning~\citep{ouyang2022training, zhou2024archer} or supervised fine-tuning~\citep{pareja2025unveiling}.

\paragraph{Impact Statement.}
This work may improve the auditability of language models by making attention patterns more sparse and easier to inspect. Sparse circuits can help identify which tokens, heads, and edges contribute to a prediction, but they should not be treated as complete explanations of model behaviour. We therefore recommend evaluating sparsity together with performance preservation and interpretability diagnostics.

\bibliographystyle{plainnat}
\bibliography{references}
\newpage
\appendix
\onecolumn
\section{Two-Digit Addition Study}
\label{app:toy_model}
\begin{figure}[h]
    \includegraphics[width=\textwidth]{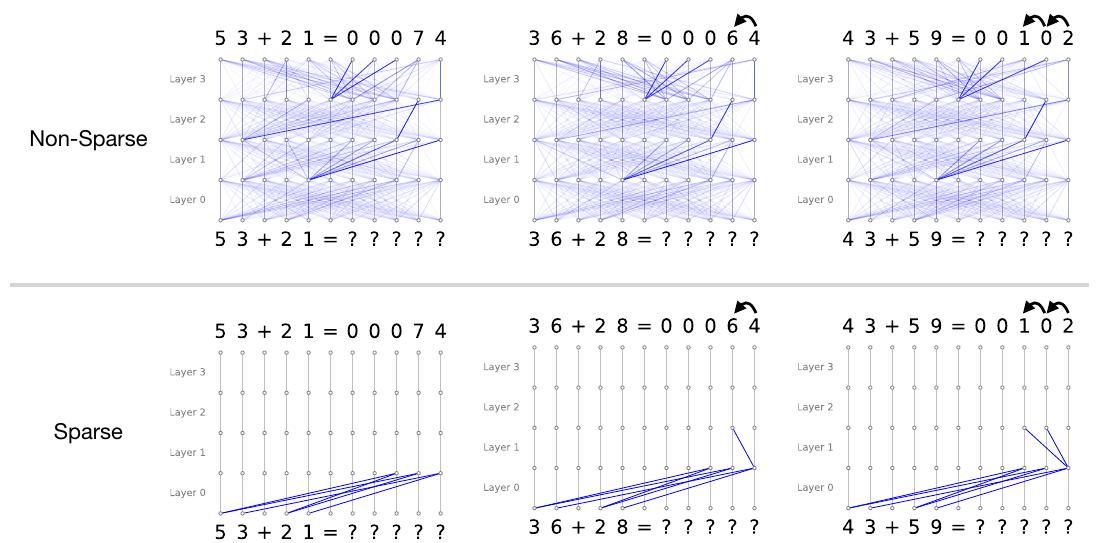}
    
    \caption{Simple example showing the attention patterns (shown in blue) of sparse and non-sparse transformers trained on a two digit addition task. Both models are able to correctly predict the sum, but the attention patterns are very different: the non-sparse model solves the task with highly dispersed information flow, while the sparse model uses a highly interpretable attention pattern: in Layer 0, the model first attends to the corresponding digits to be added, then in Layer 1, it attends to the carry bit only if it is needed (see middle and right columns, where the model has to carry once and twice respectively).}
    \label{fig:toy_model}
\end{figure}
In the introduction, we used a two-digit addition task to demonstrate how sparse attention patterns can lead to intrinsically interpretable circuits. The result presented is gathered in a small scale toy experiment described below.
We train 4-layer single-head Transformer models on a two-digit addition task, where the input is a sequence of digits and the model is trained to predict the sum. In this task, there are 13 total tokens: ten digits and three symbols "+", "=" and "?".

Within this setting, we train two models: a standard transformer model and a sparse transformer with a fixed sparsity regularisation strength.
Figure~\ref{fig:toy_model} shows several examples of the learned attention patterns.
In these examples, we can clearly see that the pressure of sparsity leads to the emergence of human-recognisable algorithmic patterns: in the first layer, each digit in the answer attends to the corresponding digits in the input, while the second layer computes the carry bit when necessary.
By enforcing selective information flow through sparse message-passing, the sparse model is able to learn crisp and localised mechanisms that are immediately amenable to interpretation.

\newpage
\clearpage
\section{Sparse Attention Implementation}
\label{app:splash_attention}

\begin{figure}[h]
    \centering
    \includegraphics[width=0.95\linewidth]{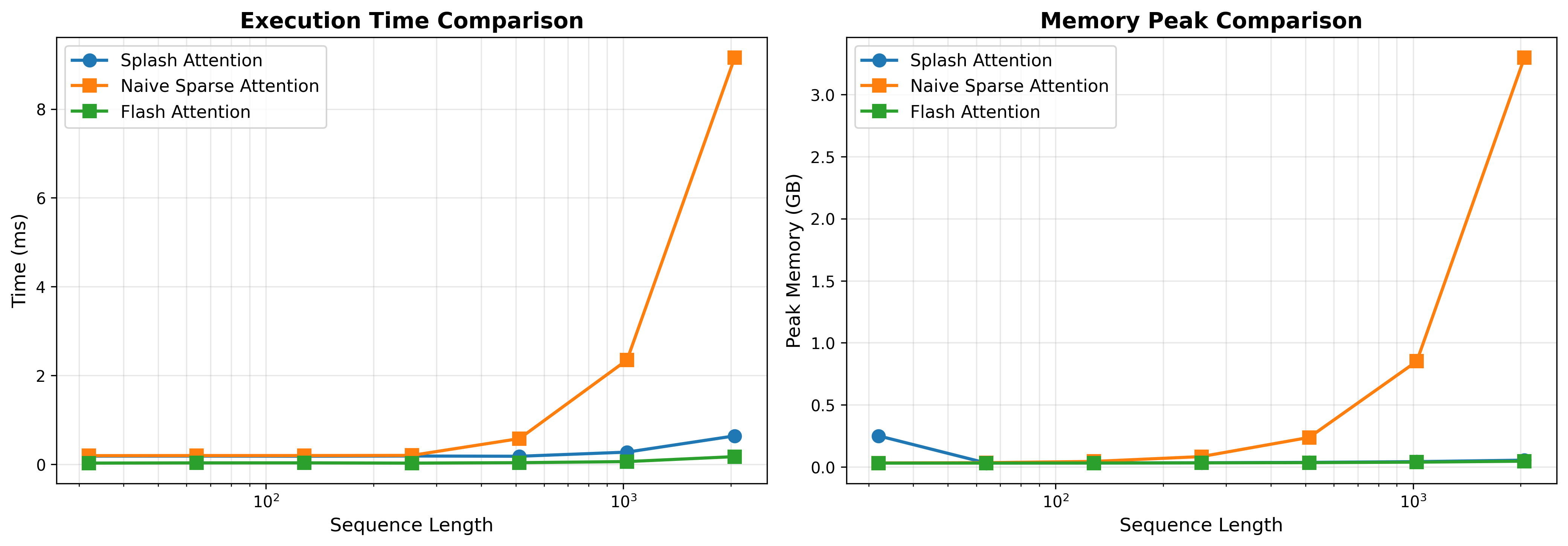}
    \caption{Performance comparison between our implementation (Splash) and a naive PyTorch sparse attention baseline. Input tensors have shape $(\texttt{batch}, \texttt{heads}, \texttt{tokens}, \texttt{head dim})=(8, 8,2^{[5,6,7,8,9,10,11]},16)$ and data type $\texttt{bfloat16}$.}
    \label{fig:splash}
\end{figure}

For the experiments, we implemented efficient GPU kernels for the sparse attention layers using the \texttt{helion} domain-specific language\footnote{\url{https://helionlang.com/}}.
We refer to this implementation as \textbf{Splash Attention} (\underline{Sp}arse f\underline{lash} Attention).
Our implementation follows the same core algorithmic structure as FlashAttention-2~\citep{dao2023flashattention}, including the use of online softmax computation and tiling.
Note that the sparse attention variant (Eq.~\ref{eqn:sparse_attn}) only differs from the standard attention by a pointwise multiplication of the adjacency matrix, which can be easily integrated into FlashAttention by computing $A_{ij}$ on-the-fly.
We additionally fuse the Gumbel-softmax computation, the straight-through gradient, and the computation of the expected number of edges (required for the penalty) into a single optimized kernel, the implementation of which will be released together with the experiment code.
Figure~\ref{fig:splash} compares our Splash Attention implementation against a naive sparse attention baseline based on PyTorch-native operations, and FlashAttention.
The resulting implementation achieves similar performance gain as FlashAttention compared to the naive attention.
Splash Attention achieves approximately 45\% MFU on an NVIDIA A100, which is slightly lower than FA, but is expected, as the gating mechanism requires additional pointwise operations.

\newpage
\section{Training Details}
\label{app:training}

\subsection{Hyperparameters and Compute Resources}

\begin{table}[h]
\centering
\begin{tabular}{lll}
\toprule
\textbf{Hyperparameter} & \textbf{OLMo} & \textbf{GPT-2}\\
\midrule
Base Model & \texttt{allenai/OLMo-7B-hf} & \texttt{gpt2}\\
Context window & 512 & 64\\
Dataset & \texttt{dolma-v1} & \texttt{OpenWebText}\\
Batch size & 16 & 256\\
Gradient accumulation steps & 4 & 4\\
Total steps & 400{,}000 & 1{,}200{,}000\\
Learning rate & $1\times10^{-5}$ & $1\times10^{-5}$\\
Minimum learning rate & $1\times10^{-6}$ & $1\times10^{-6}$\\
Optimizer & Adam & Adam \\
Weight decay & 0.1 & 0.1\\
Scheduler & Cosine (1 cycle) & Cosine (1 cycle) \\
Warmup steps & 1{,}000 & 1{,}000\\
\midrule
Finetuning strategy & LoRA & Full\\
LoRA rank ($r$) & 400 & -\\
LoRA scaling ($\alpha$) & 800 & -\\
LoRA dropout & 0 & -\\
LoRA target modules & \texttt{q,k,v,o,fc\_in,fc\_out} & - \\
\midrule
Dual Optimisation LR & 0.01 & 0.1\\
Target cross-entropy & 2.29 & 3.5\\
\bottomrule
\end{tabular}
\caption{Key hyperparameters used for sparse post-training experiments on OLMo-7B.}
\label{tab:hyperparameters}
\end{table}
We provide the key hyperparameters for our experiments in table~\ref{tab:hyperparameters}. All experiments were conducted on a single NVIDIA H100 GPU. Both GPT-2 and the OLMo model were trained in this single-GPU setting to reflect the computational constraints commonly encountered in interpretability academic research. The total training time is 7 and 14 days for GPT-2 and OLMo respectively. The main implementation code is provided as supplementary material and will be publicly released upon publication, together with the model weights.

\subsection{Training Dynamics}
\begin{figure}[h]
    \centering
    \includegraphics[width=\linewidth]{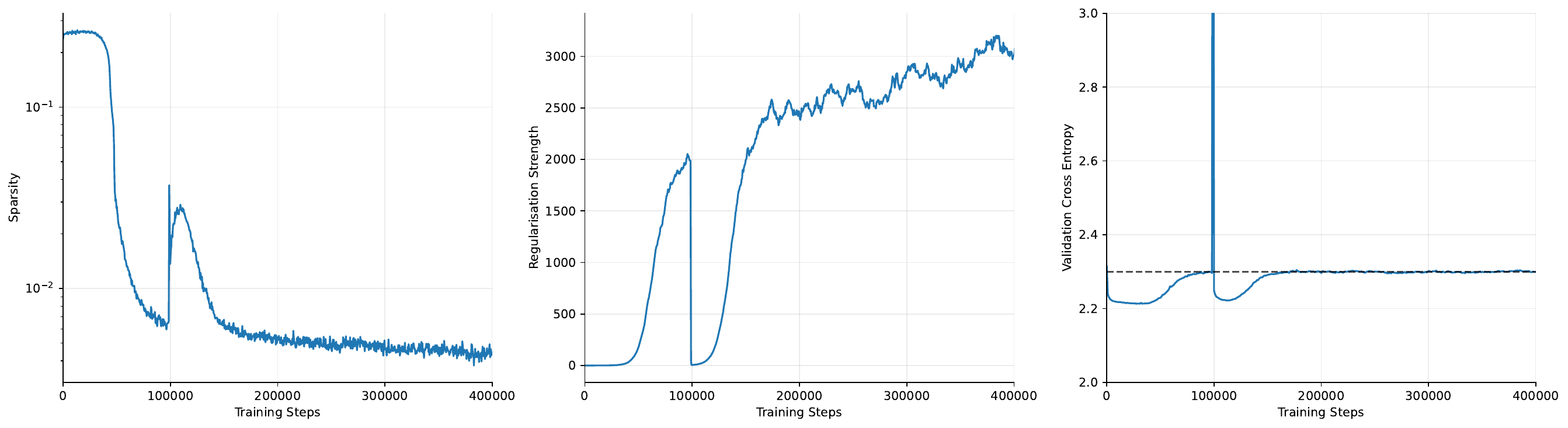}
    \caption{The training curves for post-training OLMo-7B tacking the model sparsity (left), regularisation strength (middle), and the cross-entropy loss (right). The black dotted line on the cross-entropy plot indicates the pre-defined threshold, $\tau$.}
    \label{fig:training_curve}
\end{figure}

A key feature of our post-training framework is that the strength of the sparsity regularisation is automatically controlled via a constrained optimisation scheme. By pre-specifying an accepted level for the cross-entropy target, $\tau$, the training procedure can be written as the max-min objective:
\begin{equation}
    \max_{\lambda>0}\min_\theta \bigg [ \sum_l \mathbb{E}\big[|A_l|\big] + \lambda(CE - \tau) \bigg],
\end{equation}
which can be optimised by taking alternating gradient steps in the model weight space and in the $\lambda$ space. The resulting training dynamics means that the sparsity regularisation strength increases when the model cross-entropy is lower than the target, and decreases when the model is above the threshold. Figure~\ref{fig:training_curve} shows the training curves for the OLMo-7B model. Here, we observe that the strength of sparsity regularisation keeps increasing slowly while the model cross-entropy is clipped at the desired level. Note that during a model spike (at around 100K steps), the sparsity regularisation automatically decreases to let the model recover. 

\newpage
\section{Benchmark Performance}
\label{app:benchmark}

\begin{figure}[h!]
\centering
\includegraphics[width=0.8\linewidth]{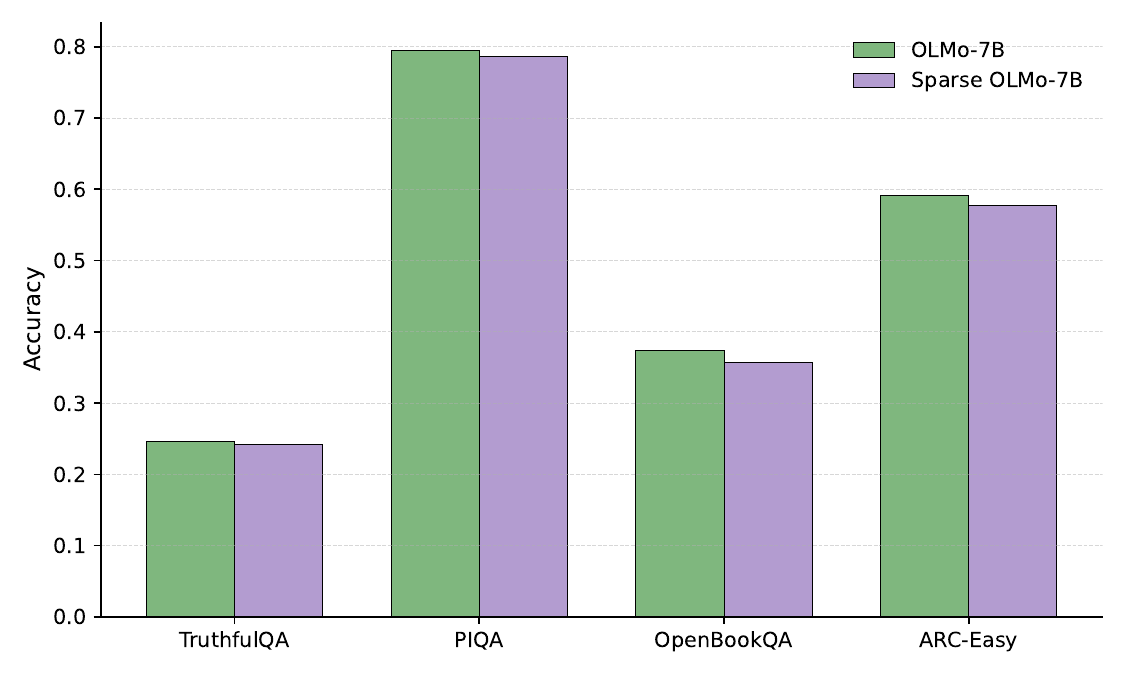}
\caption{Comparison of model performance between the base OLMo model and the sparsified model across several benchmarks.}
\label{fig:olmo_benchmark}
\end{figure}

We report additional benchmark evaluations comparing OLMo-7B with its sparsified counterpart. As shown in Figure~\ref{fig:olmo_benchmark}, the sparsified model retains performance close to the dense baseline across the evaluated tasks, despite using only a small fraction of active attention edges.

\paragraph{Long-context evaluation.}
While evaluating long-context performance of sparse-attention is a meaningful question, it is not the primary focus of this work, which is centered on mechanistic interpretability. In practice, interpretability research is typically conducted on relatively short contexts and controlled tasks (e.g., \cite{nanda2023progress, conmy2023towards}), where circuits can be precisely analysed. The complexity of the tasks considered in our work is therefore commensurate with, and in some cases exceeds, what is standard in the field, even if the absolute context length is modest. To nevertheless probe whether our method extends to longer contexts, and given that the original OLMo-7B context-window is 2048, we conduct a preliminary experiment using a LLaMA~3.2~1B model with a context window of 8192. We finetune both dense and sparse variants on Dolma 3 LongMiNo Mix for 5B tokens. The sparse model achieves approximately $0.4\%$ active attention patterns while maintaining a cross-entropy comparable to the dense model. We evaluate both models using the same context on two long-context retrieval tasks, Needle-in-a-Haystack and Passkey Retrieval \citep{liu2024lost, mohtashami2023landmark}. The dense model achieves accuracies of $0.918$ and $0.925$, respectively, while the sparse model achieves $0.919$ and $0.922$. These results show that our method can preserve long-context performance even under extreme sparsity. However, we emphasise that this experiment should be viewed as preliminary evidence rather than a comprehensive evaluation. A full investigation of interpretability in frontier long-context regimes (e.g., 64k--128k tokens) remains an open challenge for the field.

\newpage
\section{Extra Experiments for Circuit Discovery}
\label{app:extra_experiments}
In this section, we provide additional results for the activation patching circuit discovery experiment presented in the main text. 

\subsection{Sparse Circuit Visualization}

\begin{figure}[h!]
    \vspace{3pt}
    \centering
    \includegraphics[width=0.8\linewidth]{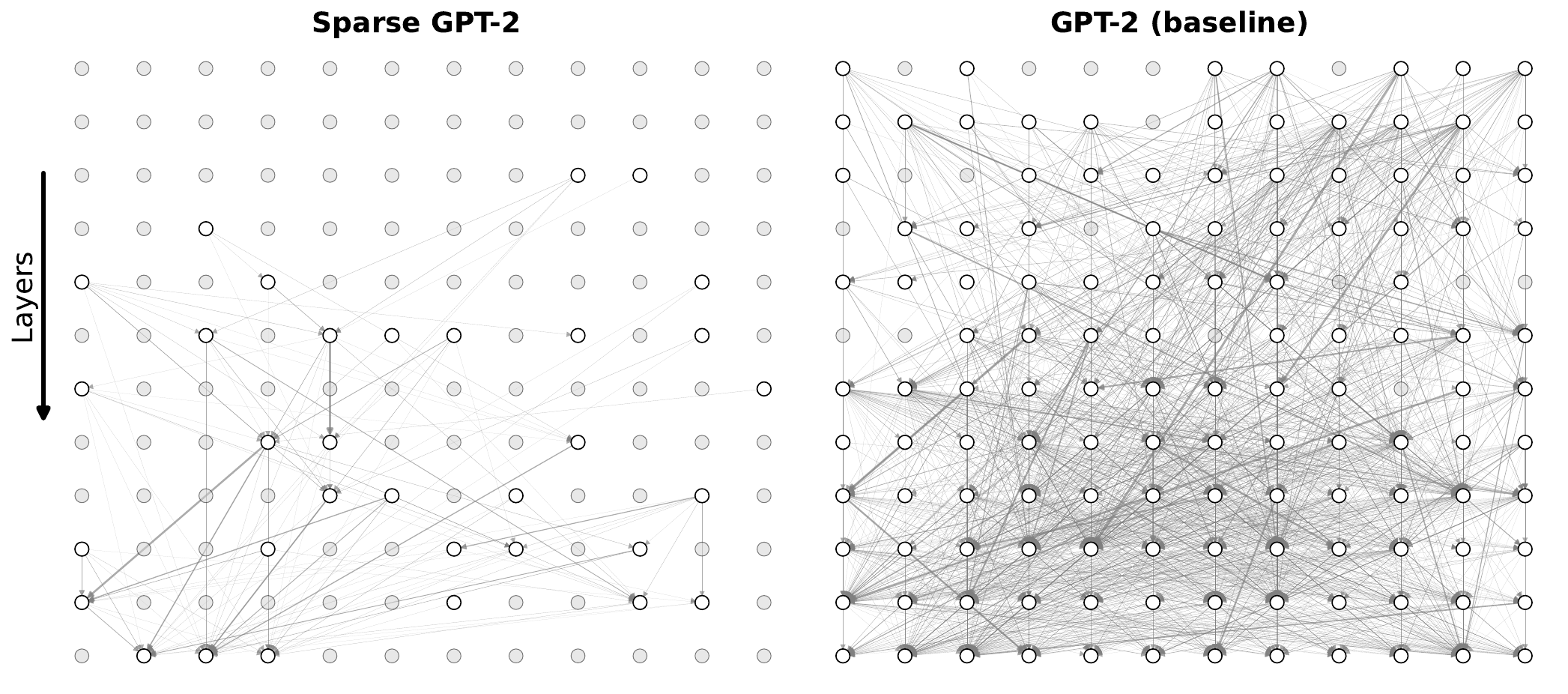}
    \caption{
    Qualitative visualization of the attention-head edges required to reach a cumulative attribution score of 0.9 on the IOI task, using scores averaged across prompts. The sparse circuit is drastically simpler, retaining only the most important task-relevant edges.
    }
    \label{fig:edge_flow}
\end{figure}

In Figure~\ref{fig:edge_flow}, we provide a qualitative visualization of the attention edges required to solve the IOI task. While Figure~\ref{fig:patching_heads_edges_single} quantifies how many edges must be preserved to maintain task performance, Figure~\ref{fig:edge_flow} shows what the resulting circuit looks like visually. The resulting sparse circuit is drastically simpler, highlighting a small subset of edges that carries most of the task-relevant computation.

\subsection{Example of Sparse-Attention Patterns}

\begin{figure}[h!]
    \vspace{5pt}
    \centering
    \includegraphics[width=0.85\linewidth]{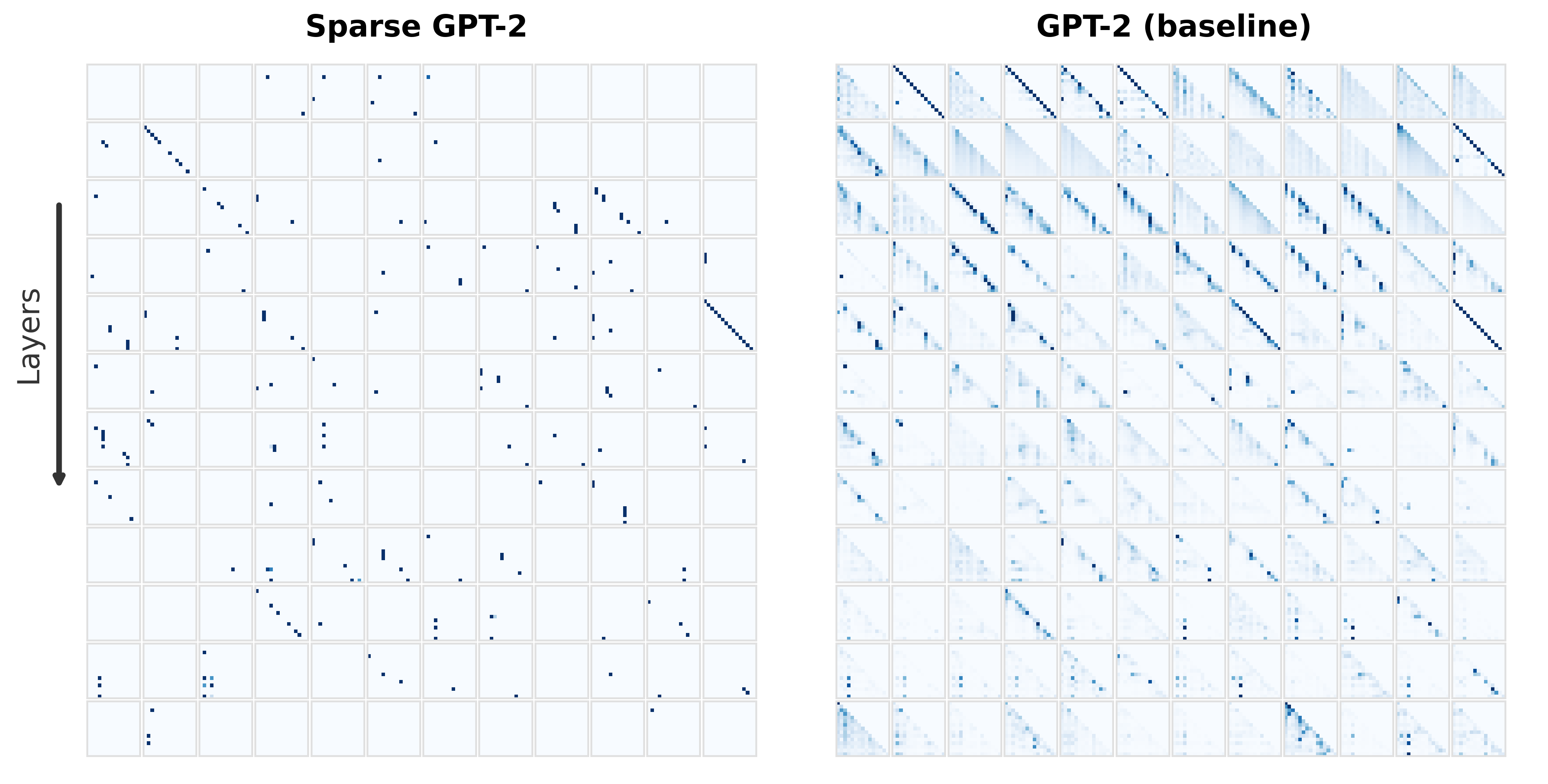}
    \caption{Attention patterns for a baseline GPT-2 model and its sparse counterpart on an IOI prompt. Darker blue entries indicate stronger attention weights.}
    \label{fig:plot_head}
\end{figure}

We illustrate an example of attention patterns across layers and heads for one instance of the IOI task: ``While John and Emma were working at the office, John gave a pen to''. Figure~\ref{fig:plot_head} compares the baseline GPT-2 model with its sparse counterpart. The sparse model exhibits a marked reduction in the number of active attention entries, with several heads appearing almost entirely inactive. Nevertheless, some heads retain structured patterns, including diagonal attention patterns.

\newpage

\subsection{Activation and Attribution Patching}

\begin{figure*}[h]
    \centering

    \vspace{30pt}

    \includegraphics[width=\textwidth]{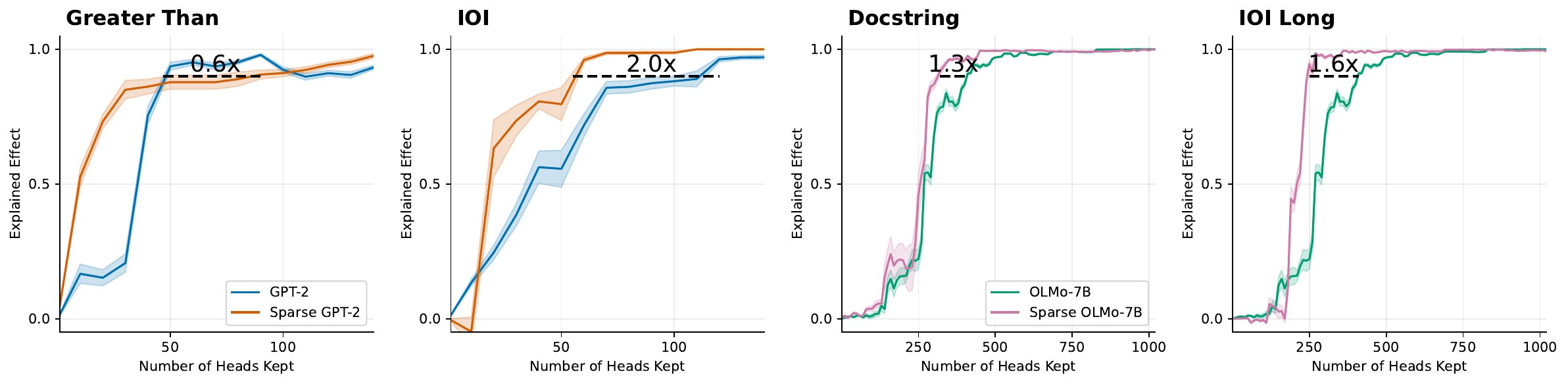}
    \vspace{0.3em}
    \includegraphics[width=\textwidth]{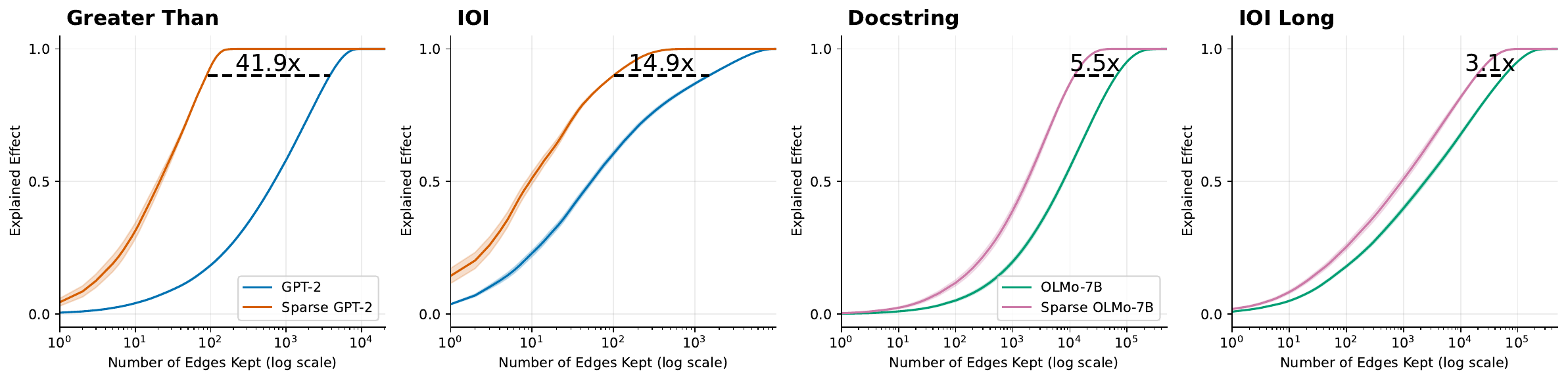}

    \caption{
    Logit attribution when keeping only the top-$k$ attention components according to a \textit{global}, task-level ranking.
    \textbf{Top:} attention \textbf{heads}. \textbf{Bottom:} attention \textbf{edges}.
    Dotted lines indicate the number of components required to explain 90\% of the logit difference.
    In contrast to the single-prompt ranking in Figure~\ref{fig:patching_heads_edges_single}, each component is assigned one importance score pooled across prompts from the same task.
    Overall, sparse models yield smaller circuits than dense models. Shaded areas show standard error across prompts.
    }
    \label{fig:patching_heads_edges_global}
\end{figure*}

Figures~\ref{fig:patching_heads_edges_global} shows the fraction of explained model preference as a function of the number of model components kept unablated. Unlike Figure~\ref{fig:patching_heads_edges_single}, where components are ranked separately for each prompt, here components are ranked at the \textit{task} level: each head or edge receives a single importance score pooled across different instances of the same task. Overall, the results are consistent with the main findings: sparse models tend to admit smaller circuits than dense models under a global ranking strategy.

\subsection{Attention Sink}
An interesting interpretability result is that sparse-attention substantially reduces persistent BOS sinks. In GPT-2, the mean fraction of attention directed to the BOS token increases after the early layers and remains high throughout later layers. By contrast, the sparse model shows BOS attention mainly in the first three layers, followed by a sharp drop after layer 2 and approximately zero attention from layer 5 onward. Results are averaged over 300 length-64 WikiText-2 sequences.

\begin{table}[h]
\centering
\caption{Mean fraction of attention directed to the BOS token across layers. Results are averaged over 50 length-64 WikiText-2 sequences.}
\vspace{5pt}
\label{tab:bos-attention}
\begin{tabular}{lcccccccccccc}
\toprule
Model & L0 & L1 & L2 & L3 & L4 & L5 & L6 & L7 & L8 & L9 & L10 & L11 \\
\midrule
Sparse & 0.24 & 0.08 & 0.10 & 0.02 & 0.06 & 0.00 & 0.00 & 0.00 & 0.00 & 0.00 & 0.00 & 0.00 \\
Dense  & 0.12 & 0.21 & 0.21 & 0.43 & 0.47 & 0.67 & 0.63 & 0.74 & 0.64 & 0.72 & 0.70 & 0.57 \\
\bottomrule
\end{tabular}
\end{table}

\newpage

\subsection{OLMo Induction Heads}

\begin{figure}[h]
    \centering
    \vspace{20pt}
    \includegraphics[width=\linewidth]{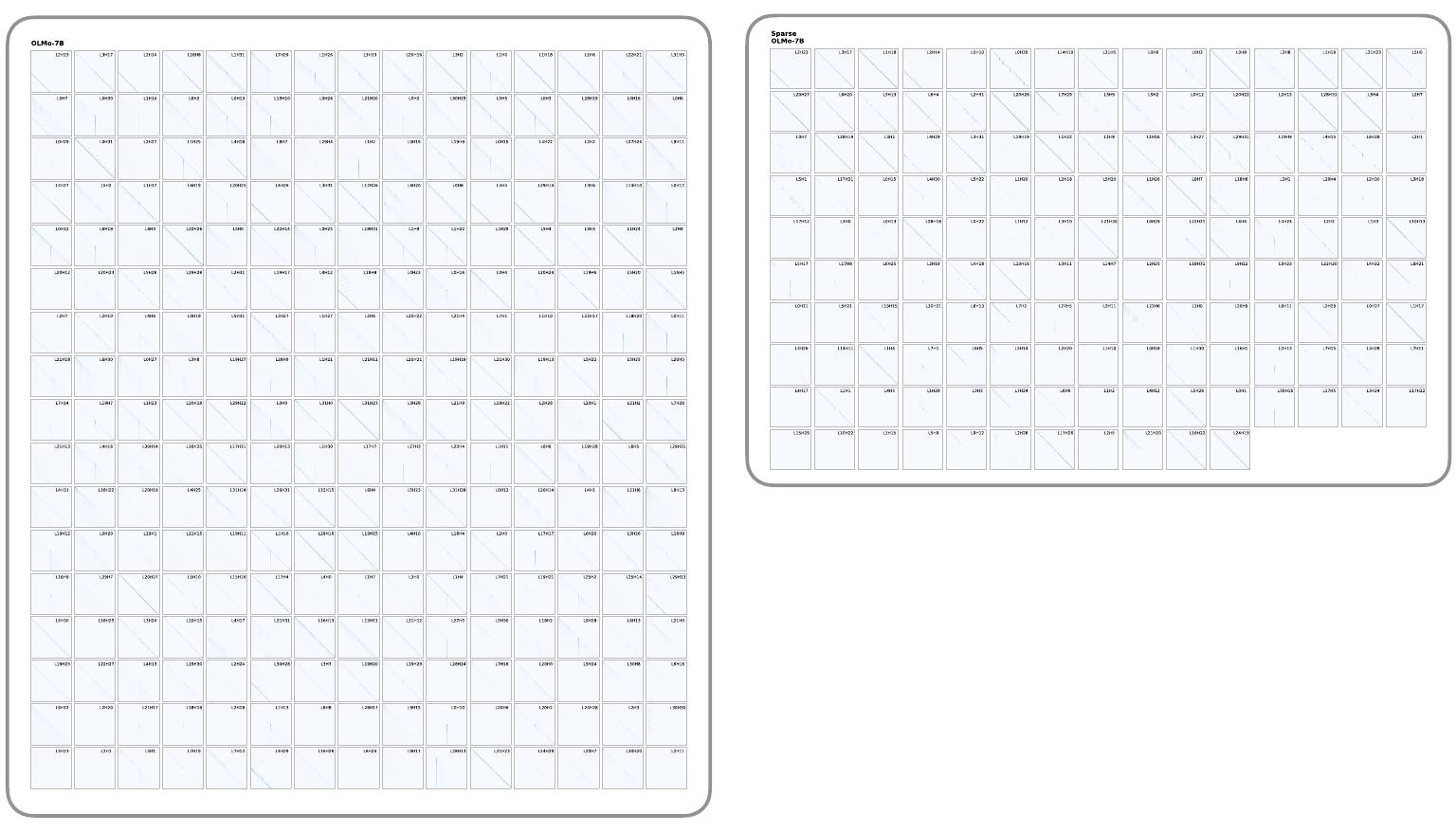}
    \caption{Attention patterns of the heads required to explain 90\% of model behaviour on a longer \textit{copy} task. Similar to the GPT-2 results in Figure~\ref{fig:induction_head}, the sparse model requires substantially fewer attention heads.}
    \label{fig:olmo_induction_head}
\end{figure}

Figure~\ref{fig:olmo_induction_head} shows the attention patterns of the heads required to explain 90\% of model behaviour on a \textit{copy} task. To fully test the longer context window afforded by OLMo, we use a longer prompt than the one used for GPT2 in the main text. The result is consistent with the GPT-2 experiment: sparsified model facilitates the discovery of smaller circuits of induction heads that implement the copy task.

\subsection{Distance Between Dense and Sparse Circuits}

Our goal is not to preserve the computational circuit implemented by the dense base model. Rather, our central claim is that dense models often implement behaviors through overly distributed circuits that are difficult to interpret. Sparse post-training is therefore intended to reorganize computation into simpler and more selective pathways while preserving model performance. From this perspective, differences between dense and sparse circuits are not a failure mode, but an expected consequence of the sparsity inductive bias.

We quantify how much the sparse circuits overlap with those of the dense model for GPT-2. On the IOI and Greater-Than tasks, we compute activation-patching importance scores for each attention head and compare the top-$k$ heads selected in the dense and sparse models. Across tasks, we observe a consistent overlap of approximately 40--50\%. This suggests that sparse post-training preserves part of the original circuit structure while also allowing the model to reorganise computation into new pathways. We also compare the models at the finer level of individual attention edges. For each sparse attention head, we match $k$ to the number of non-zero sparse edges and compare these edges with the top-$k$ attention entries in the corresponding dense head. Across 300 OpenWebText sequences, the overlap is approximately 5--10\%, compared with 2--5\% for randomly matched heads. Thus, the correspondence between dense and sparse attention patterns is limited, but reliably above chance.

Overall, these results indicate that sparse models do not simply preserve a pruned version of the dense circuit. Instead, they retain partial alignment with dense-model mechanisms while reorganising computation into sparser and more interpretable structures. This supports our view that sparse post-training induces new computational organisation rather than merely removing low-importance edges from the original model.

\newpage
\section{Circuit Discovery Tasks}
\label{app:circuit_discovery_tasks}
In the following, we provide the details and the prompts for the various tasks used in section~\ref{sec:circuit_discovery}.
\subsection{Greater-Than Task}
Each example contains a \textit{clean} prompt, a \textit{corrupt} prompt, and two disjoint sets of candidate continuations, \texttt{answers} and \texttt{wrong\_answers}. A typical entry is:

\begin{verbatim}
{
  "clean":   "The demonstrations lasted from the year 1363 to 13",
  "corrupt": "The demonstrations lasted from the year 1301 to 13",
  "answers": ["64", "65", ..., "99"],
  "wrong_answers": ["00", "01", ..., "63"]
}
\end{verbatim}

For the clean prompt, any token in \texttt{answers} yields an end year strictly greater than the start year (e.g.\ \texttt{"1364"}–\texttt{"1399"}), whereas tokens in \texttt{wrong\_answers} correspond to years that are less than or equal to the start year. The corrupt prompt changes only the starting year, shifting which continuations correspond to valid end years. We use the logit difference between the aggregated probability mass on \texttt{answers} vs.\ \texttt{wrong\_answers} in clean vs.\ corrupt contexts as our signal, in the spirit of prior mechanistic studies on simple algorithmic tasks \citep{elhage2021mathematical, nanda2023progress}.

\subsection{Indirect Object Identification (IOI) Task}

Our IOI setup follows the standard indirect object identification paradigm for mechanistic interpretability \citep{elhage2021mathematical, conmy2023towards}. Each example is generated by combining:
\begin{itemize}
    \item a pair of names \((A,B)\), e.g.\ \texttt{(" Mary", " John")};
    \item a natural-language template with placeholders \texttt{[A]}, \texttt{[B]}, and \texttt{[S]}.
\end{itemize}

We instantiate templates such as:
\begin{verbatim}
"Then, [B] and [A] went to the park. [S] gave a ball to"

"When [B] and [A] got a snack at the cafe, [S] decided to give it to"

"After the lunch, [B] and [A] went to the mall. [S] gave a gift to"
\end{verbatim}
by sampling a name pair and substituting \([A]\) and \([B]\), then choosing the subject \([S]\) (either one of the pair). The correct continuation is the indirect object, i.e.\ the other member of the pair.

For example, with \((A,B) = (\texttt{" John"}, \texttt{" Mary"})\) and \(S = B\), one instance is:
\begin{quote}
\texttt{Then, Mary and John went to the park. Mary gave a ball to}
\end{quote}
The correct continuation is \texttt{" John"}, while \texttt{" Mary"} and any distractor names are treated as incorrect candidates.

In the OLMo experiments, in order to further test the capability of our approach, we use a different set of IOI task with increased complexity and prompt length. Example templates include:
\begin{verbatim}
"After several months without any contact due to conflicting schedules and 
unexpected personal obligations, [B] and [A] finally met again at the park, 
where they spent a long afternoon catching up on past events, sharing stories, 
and reflecting on how much had changed. As the day came to an end, [S] gave 
a ball to"

"Although [B] and [A] had previously been involved in a long and emotionally
charged argument that left several issues unresolved, they agreed to meet in 
order to clarify their misunderstandings. After a tense but honest conversation, 
[S] said to"
\end{verbatim}

\subsection{Docstring Task}
We also test the OLMo models on a more complex \textit{Docstring} task~\citep{heimersheim2023circuit, conmy2023towards}, where the model needs to attend to a specific argument for a specified function in order to complete a Docstring. Similarly to the \textit{Greater Than} task, each example contains a \textit{clean} prompt, a \textit{corrupt} prompt, and two disjoint sets of candidate continuations. A typical entry is:
\begin{verbatim}
{
  "clean": "def model(self, results, old, option):
                """
                stage agency security vision spot tone joy session river unit
                :param results: bone paper selection sky
                :param old: host action hell miss
                :param",
  "corrupt": "def model(self, command, output, state):
                """
                stage agency security vision spot tone joy session river unit
                :param old: bone paper selection sky
                :param results: host action hell miss
                param",
  "answers": [" option"],
  "wrong_answers": [" results"," old"]
}
\end{verbatim}

\newpage
\section{Cross-Layer-Transcoder}
\label{app:cross-layer-transcoder}

To implement a cross-layer transcoder (CLT), let $\mathbf{h}_\ell \in \mathbb{R}^{d_{\text{model}}}$ denote the input to the MLP at layer $\ell$ for a single token position. This representation is projected into a sparse feature space via an encoder,
\begin{equation}
\mathbf{z}_{\ell} = \mathrm{ReLU}\!\left(\mathbf{W}_{\mathrm{enc}}^{\ell}\mathbf{h}_{\ell} + \mathbf{b}_{\mathrm{enc}}^{\ell}\right)
\in \mathbb{R}^{d_{\text{features}}},
\end{equation}
where $\mathbf{W}_{\mathrm{enc}}^{\ell} \in \mathbb{R}^{d_{\text{features}} \times d_{\text{model}}}$ and $\mathbf{b}_{\mathrm{enc}}^{\ell} \in \mathbb{R}^{d_{\text{features}}}$ are layer-specific encoder parameters.

The CLT reconstructs the MLP output at a target layer $\ell'$ by linearly
aggregating feature activations originating from all preceding layers,
\begin{equation}
\hat{\mathbf{m}}_{\ell'} =
\sum_{\ell \leq \ell'} \mathbf{W}_{\mathrm{dec}}^{\ell \rightarrow \ell'} \mathbf{z}_{\ell}
+ \mathbf{b}_{\mathrm{dec}}^{\ell'} ,
\end{equation}
where $\mathbf{W}_{\mathrm{dec}}^{\ell \rightarrow \ell'} \in
\mathbb{R}^{d_{\text{model}} \times d_{\text{features}}}$ denotes the decoder
mapping from layer $\ell$ to layer $\ell'$.

The summation over layers reflects the fact that a given semantic feature may
manifest in different representations across multiple MLP layers. For example,
a feature that emerges in the MLP at layer $\ell$ may reappear, potentially in a
transformed form, in the outputs of subsequent MLPs. Without accounting for
these layer-dependent variations, such duplicated representations would lead to
redundant nodes in the attribution graph. By allowing features to be represented differently across layers while being
linked through a shared latent space, the cross-layer transcoder avoids this
duplication and yields a more compact and interpretable attribution structure.
For a detailed comparison between cross-layer transcoders and standard
transcoders, we refer the reader to \citet{lindsey2025landscape}.

\begin{figure}[h]
    \centering
    \begin{subfigure}{0.43\linewidth}
        \centering
        \includegraphics[width=\linewidth]{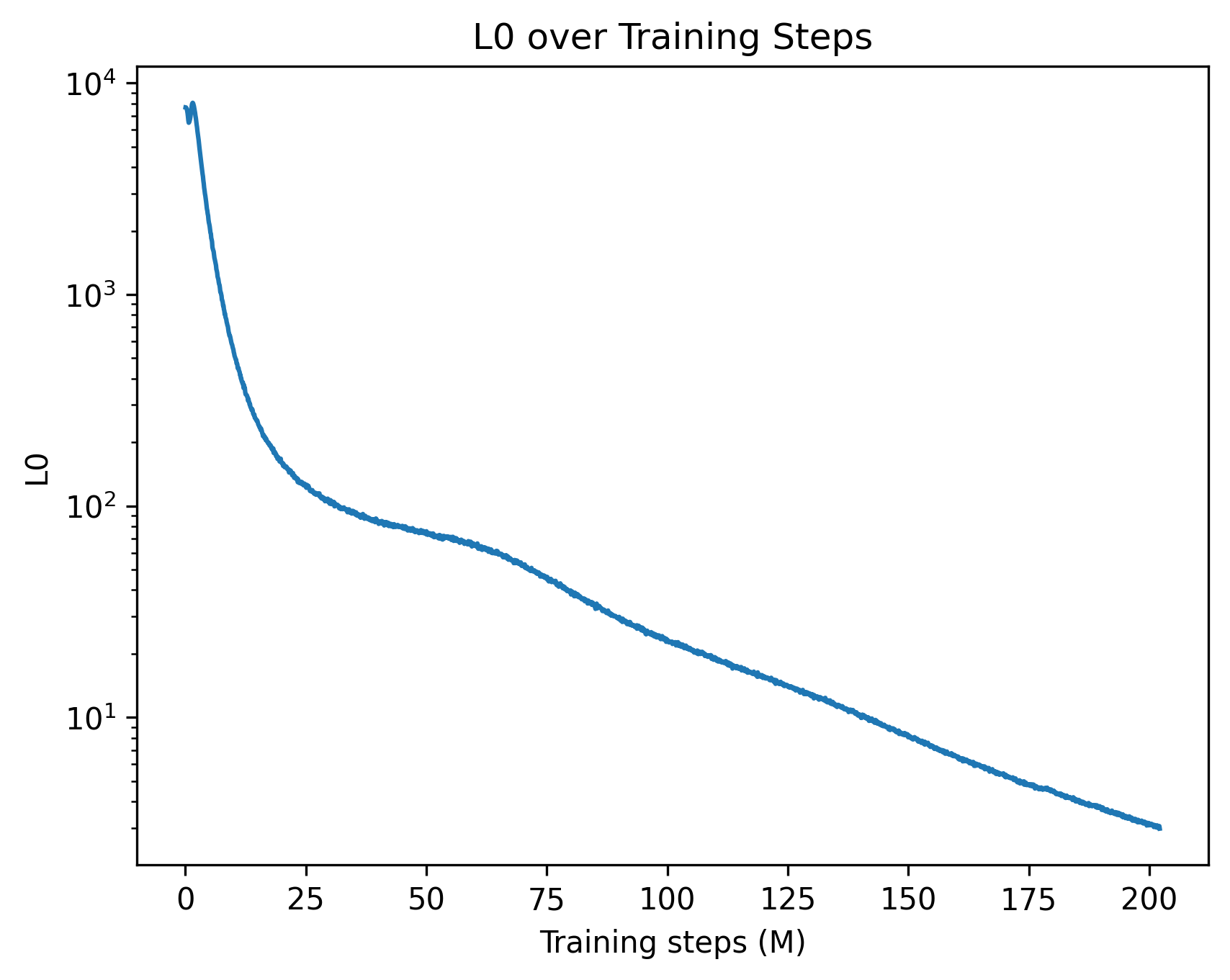}
        \caption{$L_0$ vs steps}
        \label{fig:l0_vs_steps}
    \end{subfigure}
    \hfill
    \begin{subfigure}{0.43\linewidth}
        \centering
        \includegraphics[width=\linewidth]{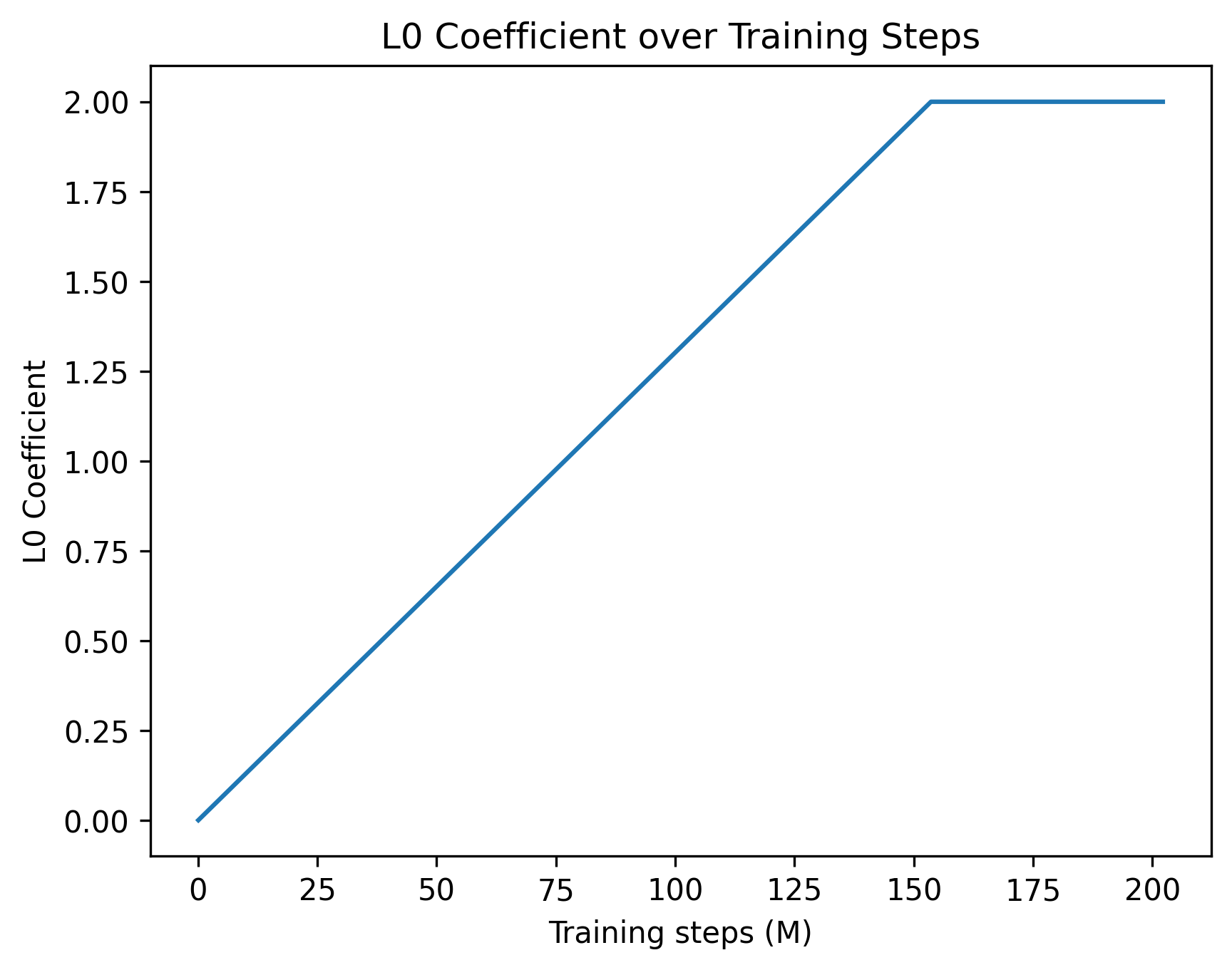}
        \caption{$L_0$ coefficient vs steps}
        \label{fig:l0_coef_vs_steps}
    \end{subfigure}

    \vspace{0.5em}

    \begin{subfigure}{0.43\linewidth}
        \centering
        \includegraphics[width=\linewidth]{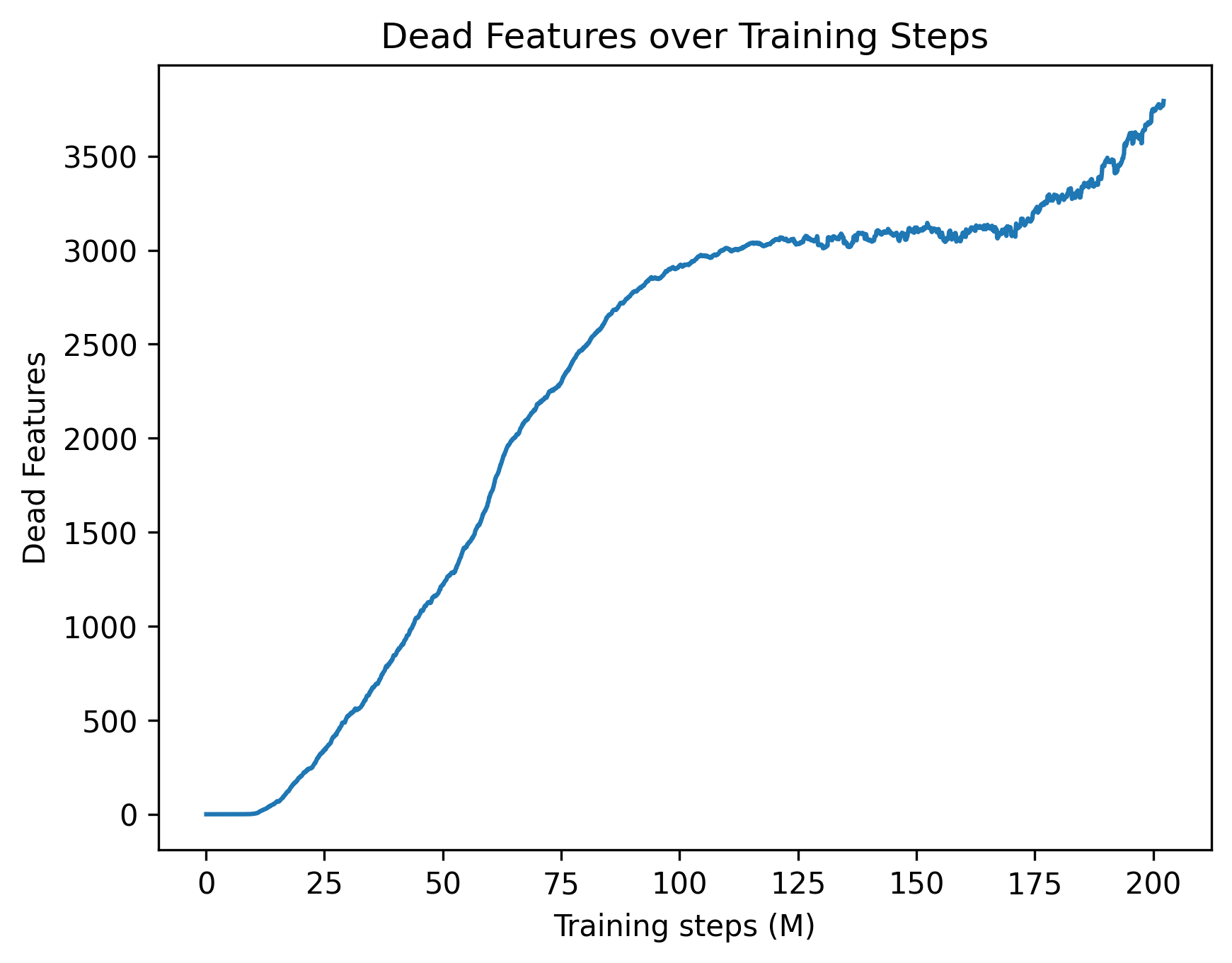}
        \caption{Dead features vs steps}
        \label{fig:dead_features_vs_steps}
    \end{subfigure}
    \hfill
    \begin{subfigure}{0.43\linewidth}
        \centering
        \includegraphics[width=\linewidth]{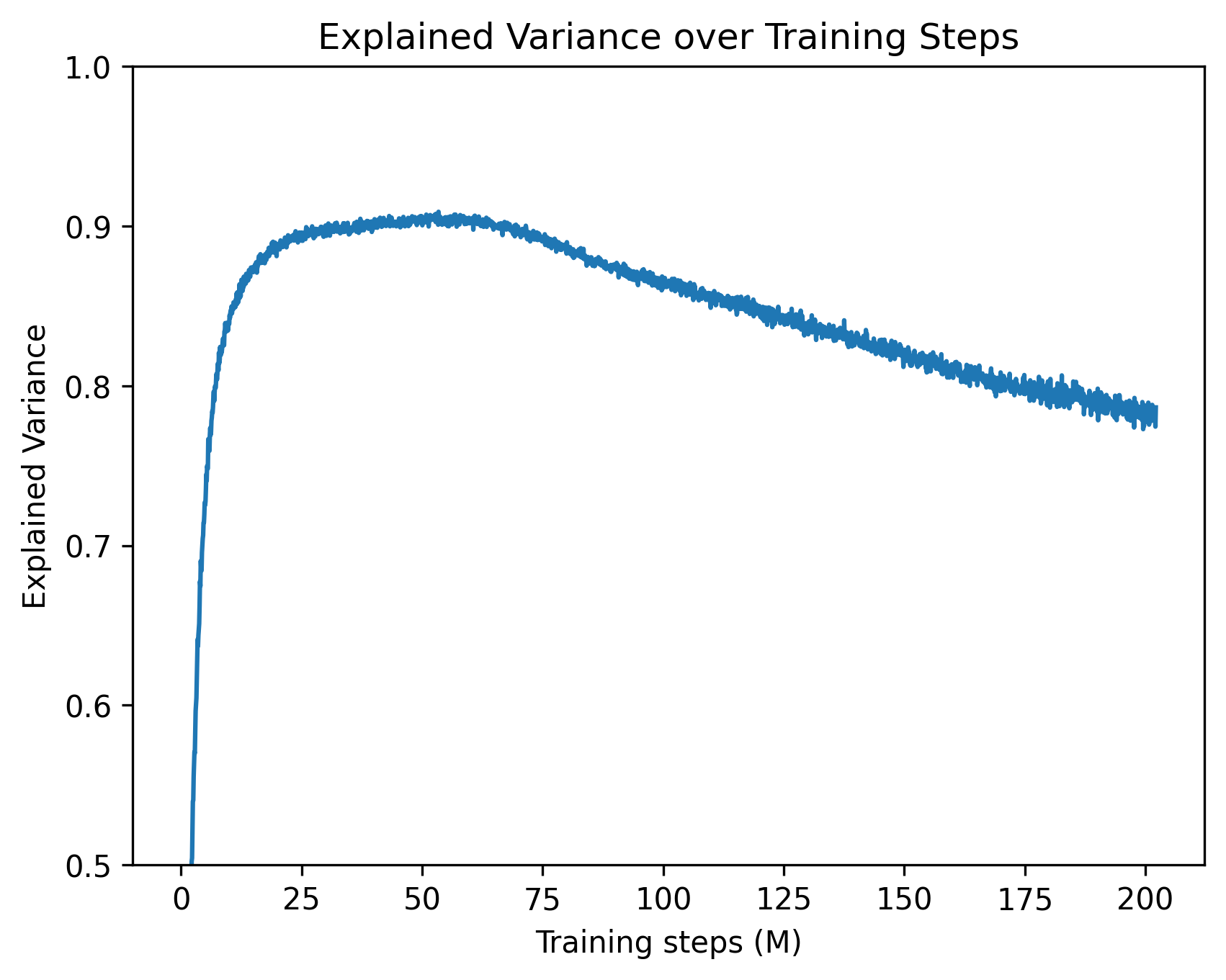}
        \caption{Explained variance vs steps}
        \label{fig:explained_variance_vs_steps}
    \end{subfigure}

    \caption{Training dynamics of the cross-layer transcoder, showing sparsity, regularization strength, dead features, and reconstruction quality over training.}
    \label{fig:training_dynamics}
\end{figure}

Following the training procedure proposed by Anthropic \citep{ameisen2025circuit}, the final objective combines reconstruction accuracy with sparsity and dead-feature regularization:
\begin{align}
\mathcal{L} =
& \underbrace{\sum_{\ell'} \left\| \hat{\mathbf{m}}_{\ell'} - \mathbf{m}_{\ell'} \right\|_2^2}_{\text{MSE reconstruction}} \nonumber \\
&+ \lambda_{0}
\underbrace{\sum_{\ell}
\tanh\!\big(C\,(\mathbf{z}_\ell \odot \|\mathbf{W}_{\mathrm{dec}}^{\ell}\|)\big)}_{\text{$L_0$ sparsity}} \nonumber \\
&+ \lambda_{\mathrm{df}}
\underbrace{\sum_{\ell}
\mathrm{ReLU}\!\Big(\exp(\tau) - \mathbf{h}_\ell^{\mathrm{pre}}\Big)
\|\mathbf{W}_{\mathrm{dec}}^{\ell}\|}_{\text{dead-feature penalty}},
\end{align}
where $\mathbf{W}_{\mathrm{dec}}^{\ell}$ denotes the concatenated decoder weights associated with layer $\ell$,
$\mathbf{h}_\ell^{\mathrm{pre}}$ are the corresponding pre-activation values,
$\tau$ is a threshold parameter, and $C$ is a scaling constant.
The hyperparameters $\lambda_{0}$ and $\lambda_{\mathrm{df}}$ control the strength of the sparsity and dead-feature regularization terms. We initialize the weights with following circuits updates \citep{transformer_circuits_jan2025}. The encoder bias is initialize to have a fixed proportion of the features active at initialization. We provide in Figure~\ref{fig:training_dynamics} the training curves of the sparsity value, the sparsity coefficient, the explained variance, and the amount of dead features. We further provide in Table~\ref{tab:training_config} the training configuration and hyperparameters used to train the GPT-2 CLTs. We hope this can help the community in training their own CLTs. 

Training CLTs is computationally expensive, with costs scaling in the number of features and layers. As a result, aside from Anthropic’s foundational work \citep{ameisen2025circuit, lindsey2025biology}, only a small number of applications exist, including a “greater-than” mechanism study \citep{merullo2025replicating}, an open-source extension \citep{lindsey2025landscape}, and a multilingual analysis \citep{harrasse2025tracing}. This scarcity highlights the difficulty of training CLTs at scale.

\begin{table}[h]
\centering
\caption{Training configuration and hyperparameters for the GPT-2 cross-layer transcoders (CLTs).}
\label{tab:training_config}
\vspace{8pt}

\begin{tabular}{ll}
\hline
\textbf{Category} & \textbf{Setting} \\
\hline
Model & GPT-2 (\texttt{HookedTransformer}) \\
Input dimension ($d_{\text{in}}$) & 768 \\
Latent dimension ($d_{\text{latent}}$) & 24\,576 \\
Expansion factor & 32 \\
Context size & 64 \\

Batch size (tokens) & 1\,024 \\
Precision & Mixed (FP32 / AMP) \\
Device & CUDA \\
Distributed training & DDP \\[0.5em]

Optimizer & Adam \\
Learning rate & $2 \times 10^{-4}$ \\
Adam $\beta_1$ / $\beta_2$ & 0.9 / 0.999 \\
Learning rate warm-up & Cosine (1\,000 steps) \\
Learning rate decay steps & 1\,874 \\
Final LR scale & 0.1 \\[0.5em]

$L_0$ coefficient & 2 \\
Optimal $L_0$ & 3 \\
$L_0$ warm-up & Linear (18\,749 steps) \\
Dead feature penalty & $10^{-5}$ \\
Dead feature window & 250 \\[0.5em]
\hline
\end{tabular}
\end{table}
\newpage
\clearpage
\section{Attribution-Graph}
\label{app:attribution}

Following \citet{ameisen2025circuit}, we define the attribution score between
feature $n$ at layer $\ell$ and position $k$, and feature $n'$ at layer $\ell'$
and position $k'$, as
\begin{equation}
a_{\ell, k, n}^{\ell', k', n'} 
= \sum_{\ell \leq s \leq \ell'}
f_{k,n}^{\ell \rightarrow s}\;
J_{s, k}^{\ell', k'}\;
g_{k',n'}^{\ell'} ,
\end{equation}
where $f_{k,n}^{\ell \rightarrow s}$ denotes the decoder vector associated with
feature $n$ projecting from layer $\ell$ to layer $s$,
$J_{s, k}^{\ell', k'}$ is the Jacobian mapping the MLP output at $(\ell, k)$ to the
MLP input at $(\ell', k')$, and
$g_{k',n'}^{\ell'}$ is the corresponding encoder feature at layer $\ell'$ and
position $k'$. The sum in this equation reflects the cross-layer mapping of the cross-layer transcoder.

The Jacobian is computed during a modified forward pass in which all nonlinear
operations, including normalization layers, attention mechanisms, and MLPs, are
frozen using stop-gradient operations.
The resulting attribution graph is pruned by retaining only those features that
cumulatively explain $80\%$ of the contribution to the final logit, and only
those edges that account for $95\%$ of the total edge-level effect.
All attribution computations are performed using the \texttt{circuit-tracer}
library \cite{circuit-tracer}. For a complete description of the attribution graph computation and pruning, we refer the user to reading \cite{ameisen2025circuit}. 

\begin{figure}[h]
    \includegraphics[width=\textwidth]{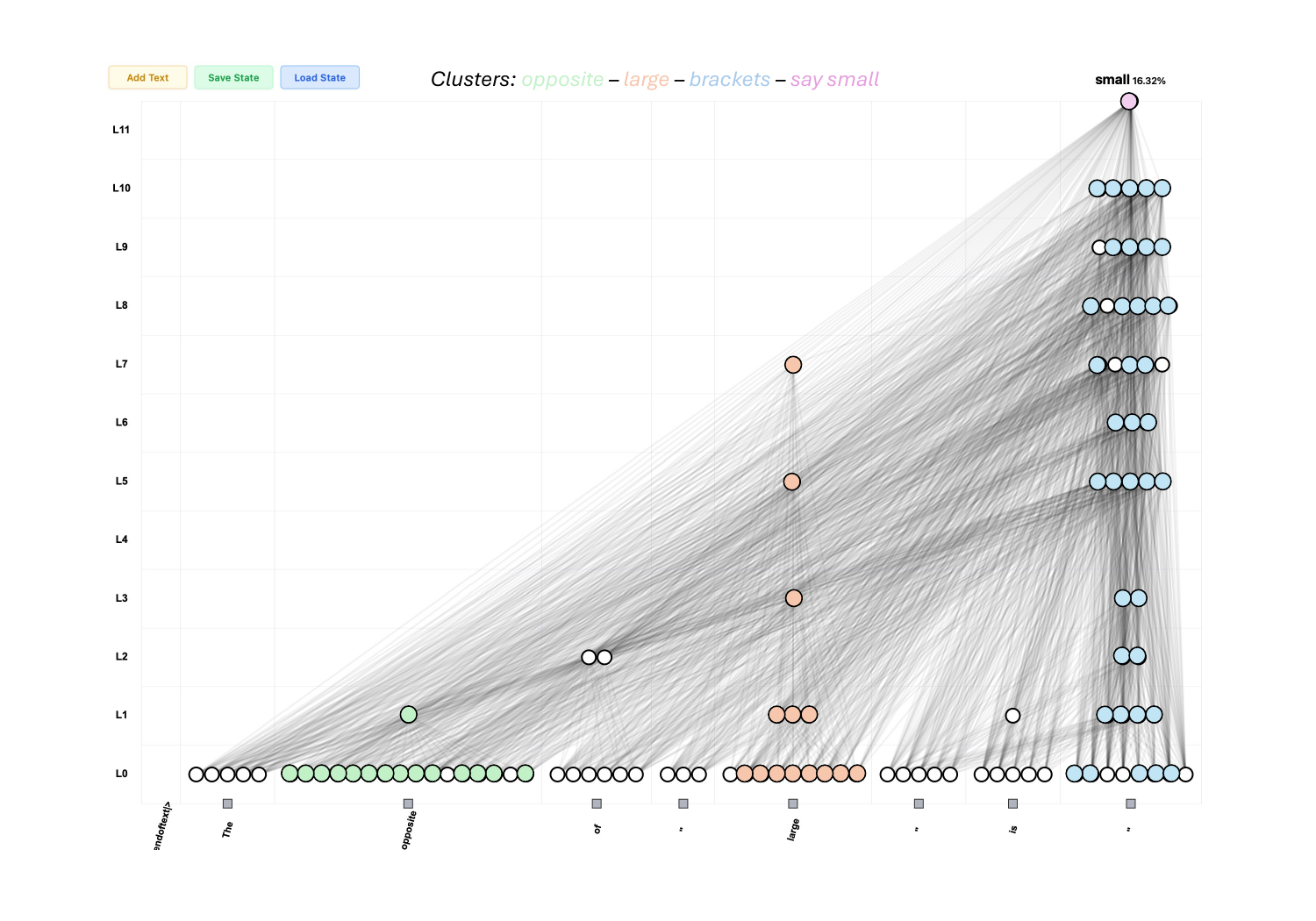}
    
    \caption{Circuit-tracing interface example for the 'The opposite of "large" is "' with GPT2-sparse.}
    \label{fig:visualization}
\end{figure}

For the visualization and the autointerp, we build our own pipeline. In Figure~\ref{fig:visualization}, we show a screenshot of the interface for the 'The opposite of "large" is "' attribution graph. The features are colored with respect to their corresponding clusters. 

\newpage
\section{Graph: The opposite of "large" is "}
\label{app:opposite_of}

We obtain a replacement score of 0.82, with 459 features identified before
pruning and 82 features remaining after pruning. The majority of features in
the resulting attribution graph fall into four dominant clusters:

\begin{itemize}
    \item \textbf{Opposition cluster}: features associated with opposition and
    comparison, primarily localized at the token position corresponding to
    \emph{opposite}.
    \item \textbf{Magnitude cluster}: features related to notions of size
    (e.g., \emph{large}, \emph{big}, \emph{full}, \emph{medium}), predominantly
    located in the residual stream at the \emph{large} token position.
    \item \textbf{Bracket cluster}: features that activate on tokens enclosed in
    brackets.
    \item \textbf{Final-logit cluster}: mainly the final logit itself and a couple of features that activate before the token "small" or related terms.
\end{itemize}

In Figure~\ref{fig:attribution_graph}, we focus on the attention heads that connect position~5 to position~8. This edge is important because it links the representation of the source cluster "large" to the representation of the target cluster "small", and therefore carries the information needed to solve the task. In the sparse attention setting, this computation is highly localized: only a small number of heads contribute substantially, and these heads attend almost exclusively from query position~8 to key position~5.

\begin{figure}[h!]
\centering

\begin{minipage}[t]{0.75\columnwidth}
\centering
\emph{$\rightarrow$ Query}\\[-2pt]
\begin{lstlisting}
1. The feature activates on tokens related to large quantities. (pos 5)
2. The feature responds again to contexts expressing large magnitude. (pos 5)
3. The feature captures references to quantities or amounts. (pos 5)
4. The feature is associated with comparative constructions. (pos 3)
5. The feature activates on tokens indicating opposition or contrast. (pos 3)
\end{lstlisting}
\end{minipage}

\caption{%
Sentence-level description of the top-5 features activating the query for the attention head L8-H6 from Figure~\ref{fig:attribution_graph}.}
\label{fig:query_features}
\end{figure}

We then ask why these heads attend to position~5. Since an attention score is determined by the interaction between the query vector at the destination position and the key vector at the source position, we decompose this interaction into the features that contribute to each side. Concretely, we identify which features at position~5, after projection through the key matrix, increase attention from position~8, and which features at position~8, after projection through the query matrix, select for those keys. This gives a feature-level explanation of why the attention edge from position~5 to position~8 appears. The computation is analogous to the causal tracing framework of~\cite{circuit-tracer}.

The resulting features are shown in Figure~\ref{fig:query_features}. We find that the key side is driven by features at position~5 that encode the relevant source-cluster information, including features related to the ``large'' and ``opposite'' clusters. This gives the following mechanistic interpretation. The model identifies the relevant opposite-cluster relation at position~5. This relation activates a small set of sparse attention heads, which route information from position~5 to position~8. The value/output pathway of these heads then linearly transforms the representation of the ``large'' cluster into the representation of the ``small'' cluster. In this sense, the attention heads do not merely correlate with the solution; they implement the routing step that allows the model to map one cluster to its opposite.

\newpage

\paragraph{Feature Descriptions.} In the boxes below, we present the top activations of representative feature
sets for each cluster.

\begin{featurebox}
{1117}
{"Opposite" cluster}

in Washington has now adopted the \hl{wider} measure of student debt outstanding. This new

the situation in Syria, Iran and the \hl{wider} region. "The

recharged by the \hl{wider} dense forests of Sanjay Van and its overflow drained

public, with interesting accounts of Oswald's demeanor at this significant moment

has a slightly \hl{wider} range. Specifically, the Atom-powered NANO

56 becoming part of the \hl{wider} Seven Years' War in which Britain and France
\end{featurebox}

\begin{featurebox}
{1337}
{"Opposite" cluster}

\hl{opposite,} piece of Mexico's cultural identity. I made the hour

\hl{opposite} shows, or something bigger, ``where there's villains

\hl{opposite} sides of Mars in 2004 and used their instruments to discover geologic evidence

\hl{opposite,} but not anymore. Now everything he says to me is some kind

\hl{opposite} direction, and had little trouble finding space at the campsites.

always seem to be just the \hl{opposite.}

show a growing trend to cast ``no" votes , \hl{opposing} how much salary

and the occupation of the \hl{opposing} forces was generally limited to mutual observation.

work hand in hand for the purpose of \hl{opposing} all movements of the thinking part

the defense's inability to stop \hl{opposing} run games. The Bills have

\hl{ing opposing} quarterbacks. The Seahawks not only had depth, they were versatile.

to win more hand battles particularly when the \hl{opposing} tackle neutralizes his initial
\end{featurebox}

\begin{featurebox}
{901}
{"Large" cluster}

Let's be honest: When someone advocates for \hl{large-scale} Muslim

robot provides a tragicomic reminder of why RWD needs to consider \hl{large} as

what kind of social safety nets should be in place to protect people from \hl{large}

advocates to limit the power of \hl{large,} established corporations, analysts say.

of \hl{large} law firms is that they are so great that the only reason anyone

that by scaling up tests, the method would be conducive for use on \hl{larger}
\end{featurebox}

\begin{featurebox}
{933}
{"Large" cluster}

people healthy and anticipating health issues before they become a problem . \hl{Big} Data is

\hl{Big} brown bucks with funny accents." Judy flinched at

\hl{BIG} UP UBUNTU: Ubuntu releases are named after industry.<|endoftext|>

\hl{BIG} LEAGUE: Barron's Says The

they need to submit their content in the same way \hl{. Big} enough

apps and offering alternatives routes \hl{. Big} data and optical fiber
\end{featurebox}

\begin{featurebox}
{1004}
{"Large" cluster}

guide said was ? \hl{full} of drinking saloons, dime museums, small

would have 2 mana sources next turn (unless his hand was \hl{full} of fast

's house, it \hl{'s full} of adventure itself.?

statement that all German Catholics had a right to \hl{full} transparency"

glimpsing a lobby \hl{full} of construction debris. The front hallway was \hl{full} of

Jokubas had recently been reading a newspaper article which was \hl{full} of
\end{featurebox}

\begin{featurebox}
{412}
{``Brackets" cluster}

group answered either ``very" or \hl{``somewhat"} attached -- except

some work colleagues. Wilcox said she found it \hl{`highly} unlikely

very rare , ``very likely ," \hl{``high} risk," she says.

ulent. Pentagon spokesman Peter Cook said the sample was ``

on of PopMatters called the album \hl{``brilliant"} and said

arlene Lowe, described him as being \hl{``one} of my biggest supporters".
\end{featurebox}

\begin{featurebox}
{518}
{``Brackets" cluster}

Kerry said Washington and Hanoi will \hl{``continue} to have differences in opinions

the United States will \hl{``take} care of it." He told reporters after the

legislation would \hl{``provide} new enforcement tools for protecting our citizens and will help

Gary Ross, said in a statement that the Air Force is currently \hl{``short}

said, and Syrian President Bashar al-Assad would \hl{``have} to go".

introduces politics into consumer policies," said Palmor, adding that it would "
\end{featurebox}


\end{document}